\documentclass{article}

\usepackage[preprint]{neurips_2026}

\usepackage[utf8]{inputenc} 
\usepackage[T1]{fontenc}    
\usepackage{hyperref}       
\usepackage{url}            
\usepackage{booktabs}       
\usepackage{amsfonts}       
\usepackage{nicefrac}       
\usepackage{microtype}      
\usepackage{tabularx}

\usepackage{graphicx}
\usepackage{tikz}
\usetikzlibrary{arrows.meta}
\usepackage{subcaption}
\usepackage{multirow}
\usepackage{hyperref}
\usepackage{amsmath}
\usepackage{cleveref}
\usepackage{todonotes}
\usepackage{wrapfig}
\usepackage{caption}
\usepackage{tcolorbox}
\tcbuselibrary{breakable,skins}
\usepackage{float}

\usepackage{fvextra} 

\usepackage{array}  
\usepackage{enumitem} 
\usepackage[normalem]{ulem} 

\definecolor{medgreen}{HTML}{008000} 
\definecolor{medred}{HTML}{CD5C5C}   
\definecolor{respblue}{HTML}{0000CD}   
\definecolor{llamaorange}{HTML}{FF8C00} 

\newcommand{\boxheader}[1]{\textbf{\uline{#1}}}

\newcommand{\sys}{{ProbMedTOD}}
\newcommand{\bnagent}{{BayesNet Agent}}
\newcommand{\polagent}{{Policy Agent}}
\newcommand{\diagagent}{{Diagnosis Agent}}

\definecolor{gaingreen}{RGB}{6,95,70}
\definecolor{lossred}{RGB}{153,27,27}

\definecolor{basecolor}{RGB}{235,245,255}   
\definecolor{bncolor}{RGB}{230,255,230}      
\definecolor{goldcolor}{RGB}{255,243,205}    
\definecolor{wrongcolor}{RGB}{255,230,230}   
\definecolor{statecolor}{RGB}{245,245,245}  

\title{\sys{}: A Bayesian Network Guided Task-Oriented Dialogue System for Patient History Taking}

\author{%
  Vishal Vivek Saley\thanks{Corresponding author -- \texttt{vishal.vivek.saley@cse.iitd.ac.in}}\\
  IIT Delhi
  \And
  Bhavesh Gurnani\\
  IIT Delhi
  \And
  Dinesh Raghu\\
  IBM Research, India
  \And
  Mausam\\
  IIT Delhi
}
\begin{document}

\maketitle

\begin{abstract}
Patient history taking is a strategic, multi-turn process that refines diagnostic hypotheses through iterative questioning, mirroring clinical reasoning where each inquiry updates beliefs over candidate diagnoses. Deploying AI in this setting is constrained by privacy, cost, and latency, making Small Language Models (SLMs) a more desirable backbone compared to larger LLMs. However, SLMs capture only the fast, intuitive ``System 1'' side of this process and lack explicit mechanisms for the deliberative ``System 2'' reasoning needed to maintain and update diagnostic uncertainty over time. Bayesian Networks (BayesNets) offer a natural complement, providing an interpretable framework for auditable probabilistic belief tracking. Yet, clinical BayesNets are difficult to construct due to their reliance on expert-curated structure, dependencies, and parameters. 

We introduce \sys{}, which automatically synthesizes a clinical BayesNet from a small number of clinical notes in three stages: ontology building, structure prediction, and parameter estimation. \sys{} then integrates the resulting BayesNet with SLMs, enabling probabilistic reasoning in multi-turn dialogues. A \bnagent{} maintains real-time posterior distributions over diagnoses, while a \polagent{} picks diagnostically informative questions. Experiments on DDXPlus and MIMIC-IV Notes show that \sys{} outperforms LLM-only and retrieval-based baselines, with especially strong gains for smaller models.
\end{abstract}

\section{Introduction}
Patient history taking is key to diagnosis \citep{engel1973interviewing, hampton1975relative, boyd2006taking}. It is a goal-driven, multi-turn dialogue to aimed identify the principal diagnosis, the condition mainly responsible for the visit \citep{cms2024icd10guidelines}. Symptoms are often vague, partial, or overlapping, forcing clinicians to reason under uncertainty. Clinicians therefore use deliberative ``System 2'' reasoning, updating hypotheses and selecting queries by clinical value \citep{garibaldi2018hypothesis}. Table~\ref{tab:motivation} illustrates this: the doctor picks the query (Fatigue) to sharpen belief about TB.

Recently, large language models (LLMs) have shown strong results on medical QA and dialogue benchmarks \citep{li2023chatdoctor, dou2023plugmed, tu2024towards, savage2024diagnostic, pal2024gemini, singhal2025toward}. However, privacy and resource constraints in real clinical settings make SLMs an attractive practical choice. SLMs excel at the fast, fluent ``System 1'' side of dialogue, but lack an explicit mechanism to maintain and update a diagnostic belief or to select questions that maximize diagnostic value \citep{guan2021bayesian, xu2024reasoning, he2024bp4er}.

\begin{table}[h]
\centering
\small
\caption{\sys{} replicates deliberative diagnostic reasoning using a BayesNet.}
\label{tab:motivation}
\newcommand{\boxtextwidth}{0.93\textwidth}
\renewcommand{\arraystretch}{1.3} 

\begin{tabular}{|m{\boxtextwidth}|}
\hline

\begin{minipage}[t]{\boxtextwidth}
    \boxheader{Context:} Patient reports a persistent \textbf{\textcolor{medgreen}{cough}} for approximately three weeks. Initially, the cough presented with white to yellowish sputum (\textbf{\textcolor{medgreen}{colored/excess sputum}}). Over the past week, the patient has developed \textbf{\textcolor{medgreen}{hemoptysis}}, estimating the amount to be about half a teaspoon. The patient denies experiencing \textbf{\textcolor{medred}{dyspnea, chest pain, wheezing, rhinorrhea, or pharyngitis}}. The patient does report \textbf{\textcolor{medred}{chest discomfort}} associated with the cough.
    \vspace{0.1cm}
\end{minipage} \\
\hline

\begin{minipage}[t]{\boxtextwidth}
    \boxheader{Gold Response:} \textcolor{respblue}{\textbf{How has your energy been? Any any fatigue?}}
    \vspace{0.1cm}
\end{minipage} \\
\hline

\begin{minipage}[t]{\boxtextwidth}
    \boxheader{Llama3 8B:} \textcolor{llamaorange}{\textbf{Do you have a headache?}}
    \vspace{0.1cm}
\end{minipage} \\
\hline

\begin{minipage}[t]{\boxtextwidth}
    \boxheader{Llama3 8B + RAG:} \textcolor{llamaorange}{\textbf{Have lost you sense of smell or taste?}}

    \textbf{Retrieved Documents:}
    \begin{enumerate}[label=\arabic*., leftmargin=0.5cm, noitemsep, topsep=1pt]
        \item \textbf{Pertusis} causes coughing, loss of consciousness, whooping cough,....
        \item \textbf{Asthma} causes chest discomfort, colored/excess sputum, dyspnea, coughing,....
        \item \textbf{Bronchitis} causes chest discomfort, coughing, fatigue,....
    \end{enumerate}
    \vspace{0.1cm}
\end{minipage} \\
\hline

\begin{minipage}[t]{\boxtextwidth}
    \boxheader{\sys{}:} \textcolor{respblue}{\textbf{Do you have chills or fever?}}

    \textbf{Bayesian Inference:}
    \begin{enumerate}[label=\arabic*., leftmargin=0.5cm, noitemsep, topsep=1pt]
        \item \textbf{Tuberculosis (89\%)} causes coughing (82.7\%), hemoptysis (36.3\%), fever (76.5\%),....
        \item \textbf{Bronchitis (5\%)} causes chest discomfort (53.4\%), coughing (86.9\%), fatigue (49.2\%),....
        \item \textbf{Pneumonia (5\%)} causes coughing (90.4\%), hemoptysis (90.4\%), chills (67.9\%)
    \end{enumerate}
    \vspace{0.1cm}
\end{minipage} \\
\hline

\end{tabular}
\end{table}

Bayesian Networks (BayesNets) have long been used for diagnosis under uncertainty \citep{shwe1991probabilistic, jaakkola1999variational, polotskaya2024bayesian}, offering explicit belief updates and transparent symptom-disease relationships. This raises a natural question: \emph{can we combine the conversational strengths of SLMs with the reliability of Bayesian diagnosis?} To answer this question, we propose our framework \sys{}, in which at each dialogue turn, a \bnagent{} extracts symptom and risk-factor states and performs real-time Bayesian inference to update diagnostic posteriors. A \polagent{} uses these posteriors to select the most informative next question. Finally, a \diagagent{} predicts the top-5 principal diagnoses from accumulated evidence.

Unfortunately, clinical BayesNets are costly to build, requiring expert-curated ontologies, manually specified structures, and carefully estimated parameters. \sys{} addresses this with an automated pipeline that constructs a BayesNet from a limited number of clinical notes in three stages: ontology construction identifies relevant diseases, symptoms, and risk factors; structure learning infers their dependencies; and parameter estimation assigns conditional probabilities directly predicted by an LLM—no expert annotation required.

In summary, our main contributions is \sys{}, which to the best of our knowledge is the first system to combine BayesNets with SLMs for medical history taking, featuring an automated BayesNet construction pipeline and probabilistic multi-turn dialogue. On high-fidelity simulations on DDXPlus and MIMIC-IV Notes, \sys{} significantly improves upon LLM-only and retrieval-based baselines, with especially strong gains for smaller models. Code and data are available at \url{https://github.com/dair-iitd/ProbMedTOD}.

\section{Related Work}

Our work broadly belongs to the line of research in which symbolic AI and/or task-specialized modules are combined with general-purpose language models for strong task performance in areas such as complex reasoning \citep{mittal-fcorebench}, knowledge-base question answering \citep{patidar-fusic,sawhney-funfusic}, relation extraction \citep{rathore-hydre}, and dialog \citep{saley-synctod}.
Within the medical domain, neuro-symbolic approaches \citep{yang2025neuro}, such as those of \citet{10385557} and \citet{hong2024argmed}, extract logical rules from medical text to support symbolic reasoning. Related work by \citet{zhang2025bayesian, babakov2025scalability} studies BayesNet structure discovery from ontology metadata. While similar in spirit, they typically do not perform probabilistic belief estimation or maintain a dialogue-conditioned belief distribution, capabilities central to \sys{}.

With the rise of LLMs, their application in medical tasks has expanded rapidly, showing strong performance in medical QA and diagnosis benchmarks \citep{chen2023meditron70b, singhal2023large, savage2024diagnostic, pal2024gemini, tu2024towards, singhal2025toward, sellergren2025medgemma}. Recent work has focused on adapting LLMs to medical dialogue settings. \citet{li2023chatdoctor,feng2026doctoragent} fine-tuned LLaMA on doctor-patient conversations to build a QA system. Other approaches, such as \citet{xu2024reasoning} and \citet{he2024bp4er}, use chain-of-thought prompting \citep{wei2022chain} to capture clinical reasoning, distilling this into smaller models. Several systems \citep{sree2024retrieval, varshney2025medprom,rose-etal-2025-meddxagent} also integrate external medical knowledge through retrieval-augmented generation (RAG) \citep{lewis2020retrieval}; for example, \citet{xu2024reasoning} uses RAG to infer likely diagnoses from dialogue context. However, none of these systems explicitly model diagnostic uncertainty. They rely on medical relationships learned during training, without probabilistic reasoning or belief updates. In contrast, \sys{} maintains a diagnostic hypothesis using a BayesNet, enabling principled updates and structured decision-making under uncertainty.

BayesNets have a long history in medical diagnosis, offering a principled way to model uncertainty and reason under incomplete information. Seminal efforts include the probabilistic reformulation of the INTERNIST-1/QMR system \citep{shwe1991probabilistic} and the use of variational inference for scalable reasoning in the QMR-DT network \citep{jaakkola1999variational}. More recent work highlights the continued relevance of BayesNets in healthcare \citep{polotskaya2024bayesian}.

Two closely related works are those by \citet{guan2021bayesian} and \citet{pavez2024hybrid}. \citet{guan2021bayesian} use a Bayesian framework to select the next symptom to inquire based on expected information gain, aiming to reduce diagnostic uncertainty. However, their system operates on structured inputs and lacks conversational capabilities. \citet{pavez2024hybrid} combine a BERT-based classifier with a Bayesian Network for mental health diagnosis, using BERT to generate a prior over diseases that is refined via probabilistic inference. While both approaches share our focus on uncertainty-aware reasoning, they do not leverage natural language dialogue. In contrast, \sys{} integrates Bayesian inference with the fluent interaction abilities of LLMs, enabling data-driven, interpretable, and conversational medical history-taking.



\section{Problem Definition}

Let $D$ be the set of $m$ candidate diseases. A TOD system for medical history taking interacts with a patient over multiple turns. At each turn $t$, the system observes the dialogue history $H_t = \{(q_1, a_1), \ldots, (q_{t-1}, a_{t-1})\}$ of prior doctor and patient utterances denoted by $q_i$ and $a_i$. It then selects the next doctor inquiry $q_t$. At the end of the dialogue, the system outputs the top-5 diagnoses.

A natural baseline is to prompt an LLM with $H_t$ and $D$, asking it to decide the most informative $q_t$. This approach relies on implicit model reasoning and lacks an explicit mechanism to treat diagnostic uncertainty. In contrast, A BayesNet over risk factors $R$, symptoms $S$, and diseases $D$ (Figure~\ref{fig:plate}) provides a complementary, structured alternative. Given an evidence state consisting of confirmed and denied symptoms and risk factors $ES =(S^+, S^-, R^+, R^-)$, it computes a diagnostic believe as the posterior $DS=P(D \mid S^+, S^-, R^+, R^-)$, enabling explicit, interpretable belief updates.

In this work, we propose \sys{}, the first TOD system that combines conversation capabilities of LLMs with probabilistic framework of BayesNets for medical history taking. A key challenge is enabling effective interaction between the LLM and the BayesNet during dialogue to extract evidence and update beliefs. Further, clinical BayesNets are difficult to construct, requiring expert-defined ontologies, graph structures, and parameters. We address this by automatically constructing a clinically valid BayesNet from a small corpus $\mathcal{P}$ of patient profiles.

\section{\sys{} System}

\begin{wrapfigure}[27]{r}{0.48\textwidth}
        \vspace{-20pt}
        \centering
        \includegraphics[width=0.48\textwidth]{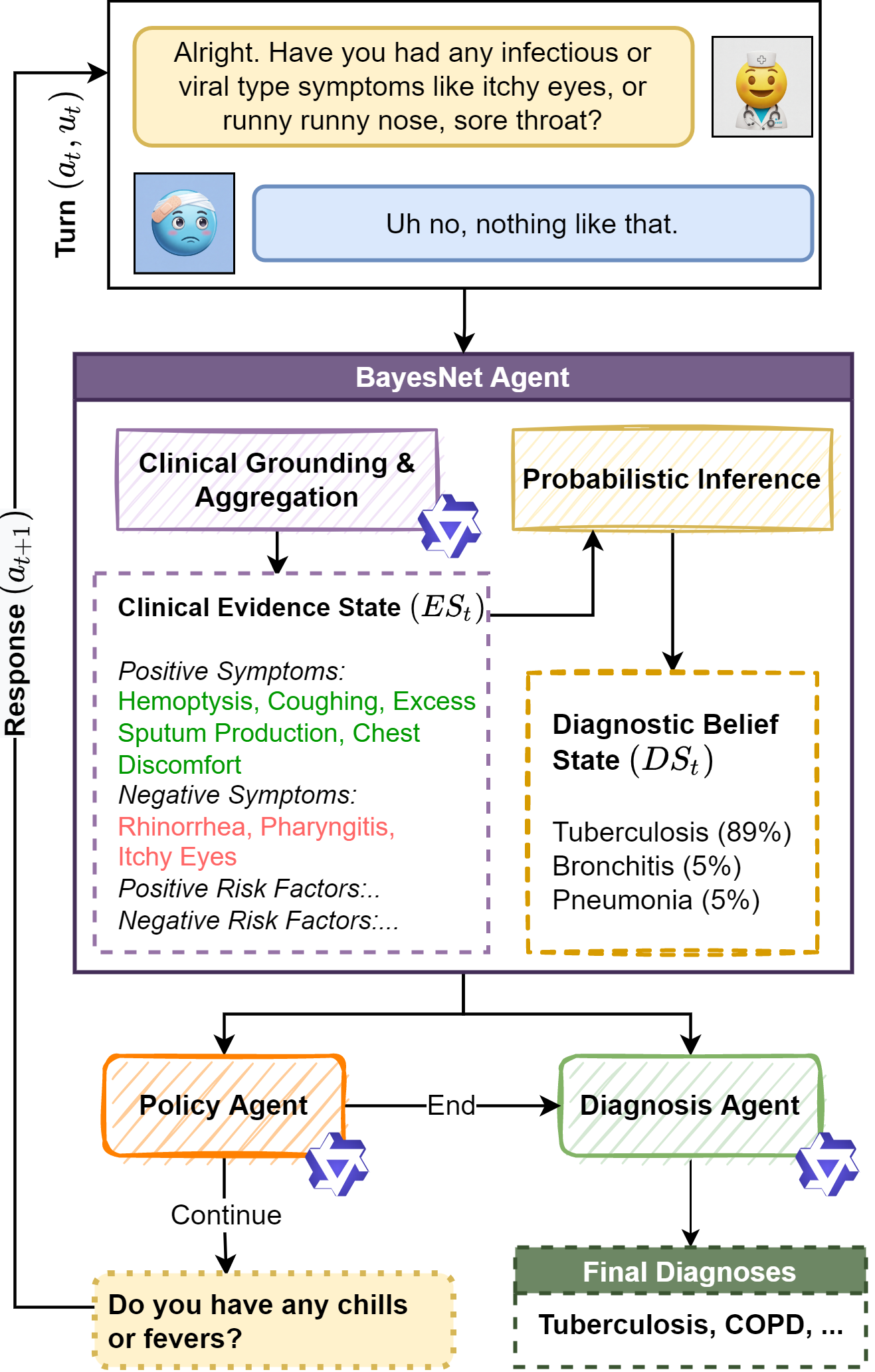}
        \caption{\sys{} architecture. Extracted Evidence state is used to compute diagnostic belief via a BayesNet.}
        \label{fig:bntod}
\end{wrapfigure}

We first describe the overall architecture of \sys{} for medical history taking assuming that the BayesNet is given. Figure~\ref{fig:bntod} provides a high-level overview of the system. \sys{} consists of three main components: \bnagent{} for evidence extraction and belief update, \polagent{} for dialogue policy, and \diagagent{} for final diagnosis. All three agents are prompt-based and use chain-of-thought (CoT) prompting. 

\paragraph{BayesNet Agent:}
At turn $t$, \bnagent{} processes the latest doctor-patient exchange $(a_t, u_t)$ to update the evidence state $ES_t$ and compute the diagnostic belief $DS_t$. A key challenge is that SLMs struggle to directly compare free-form patient utterances against large clinical ontologies. Ontologies often contain hundreds of expert terms, while patients describe symptoms in lay language (e.g., "can't catch my breath" $\rightarrow$ \emph{dyspnea}). To bridge this gap, \bnagent{} employs a retrieve–rerank pipeline that narrows the search space while leveraging the LLM's reasoning capabilities.

First, an LLM classifies whether $u_t$ contains evidence about symptoms, risk factors, or neither. This lightweight routing step avoids unnecessary retrieval when the patient response is non-clinical (e.g., clarifications or pleasantries). Second, based on the classification, candidates are retrieved separately from the symptom set $S$ and the risk-factor set $R$ using the SapBERT \citep{liu-etal-2021-self} clinical retriever. The top-10 symptom nodes and top-10 risk-factor nodes are selected, skipping irrelevant retrievals as needed.

Finally, another LLM prompt resolves each candidate as confirmed, denied, or unmentioned, yielding per-turn evidences. These findings are aggregated into the cumulative evidence state $ES_t$. Finally, \bnagent{} uses the BayesNet to compute the diagnostic belief state $DS_t = P(D \mid ES_t)$, a posterior distribution over diseases.

\paragraph{\polagent{}.} The \polagent{} has two primary responsibilities: determining whether sufficient evidence has been gathered to make a diagnosis, and formulating the next inquiry to maximize diagnostic accuracy. To achieve this, the \polagent{} combines fast, heuristic-like System 1 reasoning with the principled, probabilistic beliefs of System 2 reasoning derived from the BayesNet.

Reasoning is crucial for the \polagent{} as it allows the agent to use information outside the BayesNet ontology and course-correct when the BayesNet belief $DS_t$ is misaligned with the dialogue history $H_t$. The agent achieves this by prompting an LLM to reason through the evidence and dialogue history. If the \polagent{} concludes that enough evidence is available, it terminates the dialogue and invokes the \diagagent{}. Otherwise, it selects a follow-up question $a_{t+1}$ targeting the leading source of diagnostic uncertainty in $DS_t$.

\paragraph{Diagnosis Agent}
When the policy concludes the interview at turn $T$, \diagagent{} receives the full history $H_T$ and the final belief $DS_T$. Using the same reasoning  pattern as \polagent{}, it weighs $DS_T$ against the clinical evidence present in $H_T$, combining probabilistic and linguistic evidence to produce a ranked top-5 list of diagnoses from the candidate set $\mathcal{D}$.

\section{The BayesNet Construction}

\begin{figure}[]
    \centering
\begingroup%
\definecolor{bnInk}{HTML}{20364B}%
\definecolor{bnMuted}{HTML}{526579}%
\definecolor{bnLine}{HTML}{C5D1DC}%
\definecolor{bnBlue}{HTML}{356B9A}%
\definecolor{bnBlueFill}{HTML}{F0F5FA}%
\definecolor{bnTeal}{HTML}{287F78}%
\definecolor{bnTealFill}{HTML}{EFF7F5}%
\definecolor{bnPurple}{HTML}{756298}%
\definecolor{bnPurpleFill}{HTML}{F5F2F9}%
\definecolor{bnGold}{HTML}{A66C29}%
\definecolor{bnGoldFill}{HTML}{FCF6EC}%
\resizebox{\linewidth}{!}{%
\begin{tikzpicture}[
 x=1cm,y=1cm,text=bnInk,
 font=\sffamily\fontsize{7.5}{9.5}\selectfont,
 bn text/.style={align=left,inner sep=0pt,anchor=west,
   font=\sffamily\fontsize{7.3}{9}\selectfont},
 bn title/.style={anchor=west,align=left,inner sep=0pt,
   font=\sffamily\bfseries\fontsize{8}{10}\selectfont},
 bn badge/.style={circle,minimum size=0.32cm,inner sep=0pt,text=white,
   font=\sffamily\bfseries\fontsize{7}{8}\selectfont},
 bn small/.style={font=\sffamily\fontsize{7}{8.5}\selectfont},
 bn flow/.style={-{Stealth[length=1.8mm,width=1.3mm]},draw=bnMuted,line width=0.7pt},
 bn edge/.style={-{Stealth[length=1.5mm,width=1.1mm]},draw=bnMuted,line width=0.65pt},
 bn concept/.style={rounded corners=2.5pt,align=center,inner sep=3pt,
   minimum height=0.48cm,line width=0.6pt,
   font=\sffamily\fontsize{7}{8.5}\selectfont},
 bn risk/.style={bn concept,draw=bnTeal!55,fill=white,text=bnTeal},
 bn symptom/.style={bn concept,draw=bnBlue!55,fill=white,text=bnBlue},
 bn disease/.style={bn concept,draw=bnPurple!55,fill=white,text=bnPurple,
   font=\sffamily\bfseries\fontsize{7.5}{9}\selectfont},
 bn weight/.style={inner sep=1pt,text=bnGold,
   font=\sffamily\bfseries\fontsize{7}{8.5}\selectfont},
 bn sublabel/.style={align=center,inner sep=0pt,
   font=\sffamily\fontsize{7}{8.5}\selectfont}
]

\path[draw=bnLine,fill=white,rounded corners=3pt,line width=0.6pt]
 (0,0.45) rectangle (15.95,1.45);
\node[align=left,anchor=west,inner sep=0pt,text width=2.65cm,
 font=\sffamily\bfseries\fontsize{8}{10}\selectfont] at (0.23,0.95)
 {Patient notes\\{\normalfont\sffamily\fontsize{7}{8.5}\selectfont Synthetic example}};
\node[bn text,text width=12.63cm] at (3.10,1.18)
 {\textbf{A:} Wheezing, shortness of breath; mother has asthma; feels unwell.};
\node[bn text,text width=12.63cm] at (3.10,0.72)
 {\textbf{B:} Wheeze, dyspnea; father has asthma; recent overseas travel.};
\draw[bn flow] (2.05,0.45) -- (2.05,0);

\path[draw=bnTeal!45,fill=bnTealFill,rounded corners=4pt,line width=0.6pt]
 (0,0) rectangle (7.00,-5.55);
\path[draw=bnPurple!45,fill=bnPurpleFill,rounded corners=4pt,line width=0.6pt]
 (7.40,0) rectangle (11.45,-5.55);
\path[draw=bnGold!45,fill=bnGoldFill,rounded corners=4pt,line width=0.6pt]
 (11.85,0) rectangle (15.95,-5.55);

\node[bn badge,fill=bnTeal] at (0.26,-0.40) {1};
\node[bn title] at (0.53,-0.40) {Ontology building};
\node[bn badge,fill=bnPurple] at (7.66,-0.40) {2};
\node[bn title] at (7.93,-0.40) {Structure prediction};
\node[bn badge,fill=bnGold] at (12.11,-0.40) {3};
\node[bn title] at (12.38,-0.40) {Parameter estimation};
\draw[draw=bnTeal!25,line width=0.5pt] (0.18,-0.77) -- (6.82,-0.77);
\draw[draw=bnPurple!25,line width=0.5pt] (7.58,-0.77) -- (11.27,-0.77);
\draw[draw=bnGold!25,line width=0.5pt] (12.03,-0.77) -- (15.77,-0.77);
\draw[bn flow] (7.00,-2.85) -- (7.40,-2.85);
\draw[bn flow] (11.45,-2.85) -- (11.85,-2.85);

\node[bn sublabel,text=bnBlue,text width=2.85cm] at (1.67,-1.13)
 {\textbf{LLM extraction}};
\node[bn sublabel,text=bnTeal,text width=3.15cm] at (5.18,-1.13)
 {\textbf{SapBERT grouping + LLM decisions}};

\node[bn text,text=bnBlue,text width=2.75cm] at (0.25,-1.50) {\textbf{Symptoms}};
\node[bn text,text width=2.85cm] at (0.25,-2.28)
 {wheezing; wheeze\\shortness of\\breath; dyspnea;\\feeling unwell\\};
\node[bn text,text=bnTeal,text width=2.75cm] at (0.25,-3.10) {\textbf{Risk factors}};
\node[bn text,text width=2.85cm] at (0.25,-3.68)
 {mother/father has asthma\\recent overseas travel};

\draw[bn flow] (2.50,-2.72) -- (3.43,-2.72);
\node[bn symptom,text width=1.27cm] at (4.30,-2.00) {Wheezing};
\node[bn symptom,text width=1.27cm] at (6.00,-2.00) {Dyspnea};
\node[bn risk,text width=1.55cm] at (4.38,-3.02) {Family history\\of asthma};
\node[bn risk,text width=1.25cm] at (6.03,-3.02) {Recent\\travel};
\node[bn sublabel,text=bnMuted,text width=3.10cm] at (5.18,-3.83)
 {Removed: \emph{feeling unwell}};
\node[bn sublabel,text=bnTeal,text width=6.20cm] at (3.50,-4.91)
 {SapBERT narrows comparisons;\\the LLM merges and filters.};

\foreach \cx/\prefix in {9.425/s,13.900/e} {
 \node[bn risk,text width=1.45cm] (\prefix Family) at (\cx-0.82,-2.30)
   {Family history\\of asthma};
 \node[bn risk,text width=1.15cm] (\prefix Travel) at (\cx+0.92,-2.30)
   {Recent\\travel};
 \node[bn disease,text width=1.05cm] (\prefix Asthma) at (\cx,-3.40) {Asthma};
 \node[bn symptom,text width=1.15cm] (\prefix Wheeze) at (\cx-0.82,-4.47) {Wheezing};
 \node[bn symptom,text width=1.15cm] (\prefix Dyspnea) at (\cx+0.82,-4.47) {Dyspnea};
}

\node[bn sublabel,text=bnPurple,text width=3.55cm] at (9.425,-1.13)
 {\textbf{CDC asthma reference}};
\node[bn sublabel,text=bnMuted,text width=3.55cm] at (9.425,-1.48)
 {LLM maps evidence to nodes};
\draw[bn edge] (sFamily.south) -- (sAsthma.north);
\draw[bn edge] (sAsthma.south) -- (sWheeze.north);
\draw[bn edge] (sAsthma.south) -- (sDyspnea.north);
\node[bn small,text=bnMuted,inner sep=0pt] at (10.36,-2.92) {No edge};
\node[bn sublabel,text=bnPurple,text width=3.55cm] at (9.425,-5.16)
 {The reference supports the retained edges.};

\node[bn sublabel,text=bnGold,text width=3.60cm] at (13.900,-1.13)
 {\textbf{LLM estimates parameters}};
\node[bn sublabel,text=bnMuted,text width=3.60cm] at (13.900,-1.48)
 {$p_{\mathrm{asthma}}=0.05$};
\draw[bn edge] (eFamily.south) -- (eAsthma.north);
\draw[bn edge] (eAsthma.south) -- (eWheeze.north);
\draw[bn edge] (eAsthma.south) -- (eDyspnea.north);
\node[bn small,text=bnMuted,inner sep=0pt] at (14.82,-2.92) {No estimate};
\node[bn weight] at (13.03,-2.95) {0.20};
\node[bn weight] at (13.08,-3.92) {0.70};
\node[bn weight] at (14.72,-3.92) {0.60};
\node[bn sublabel,text=bnGold,text width=3.60cm] at (13.900,-5.16)
 {Only retained edges receive strengths.};

\end{tikzpicture}%
}%
\endgroup%
    \caption{The three stages of BayesNet construction.}
    \label{fig:bayesnet-construction}
\end{figure}
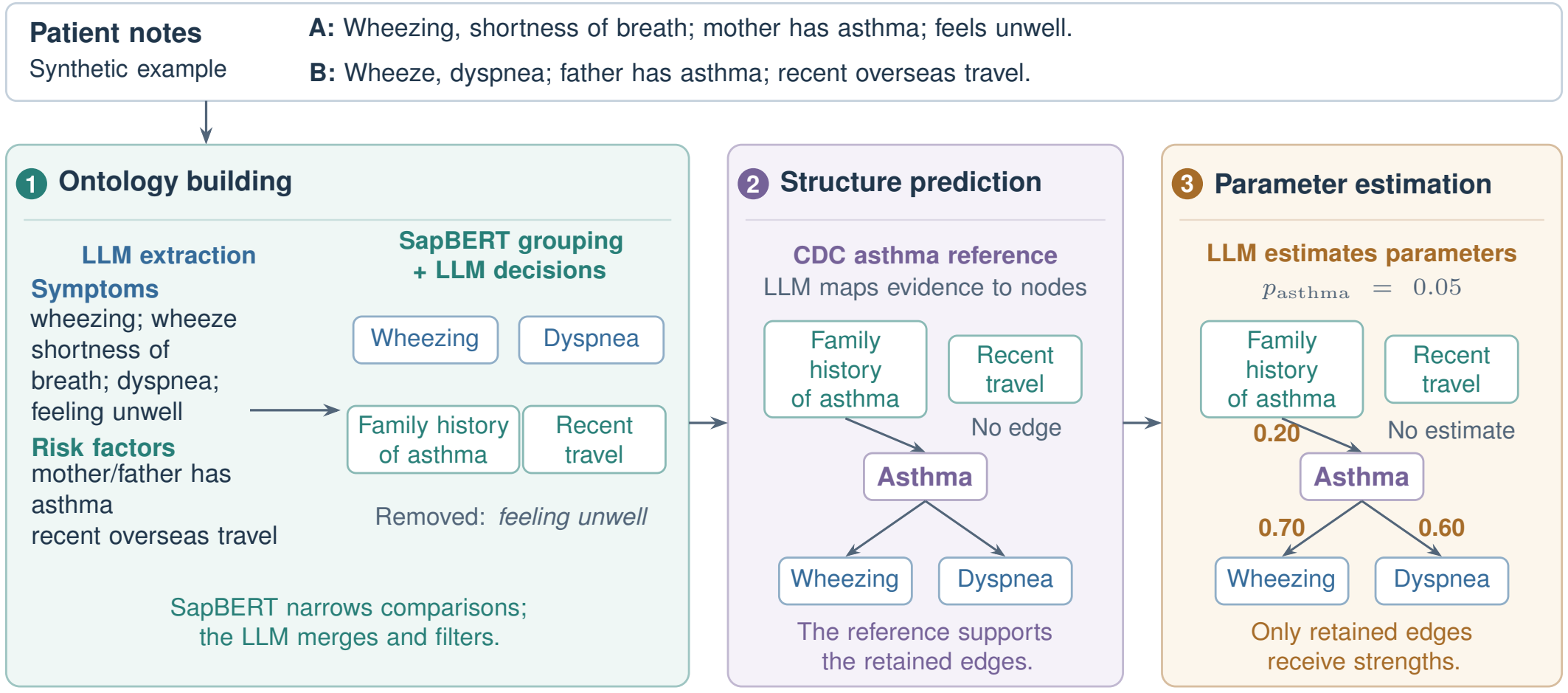

Clinical BayesNets are not always available, and constructing one from scratch is difficult. Defining the node set requires domain expertise because relevant symptoms and risk factors may be missing or inconsistently named across standard ontologies.
Specifying the graph structure demands that clinicians reason explicitly about causal relationships among diseases, risk factors, and symptoms.
Estimating the parameters is harder still: the strength of each relation must be quantified, but the patient-level data needed for statistical estimation are often costly to obtain. A BayesNet should also capture patterns in the target dataset, such as occupational or travel-related risk factors that may not appear in a standard ontology.
To address these challenges, \sys{} constructs a BayesNet from a small corpus $\mathcal{P}$ of patient notes for a given disease set $D$. Construction proceeds in three stages: \emph{ontology building}, \emph{structure prediction}, and \emph{parameter estimation}. The notes identify the symptoms and risk factors that should be represented, medical references provide evidence for the graph structure, and an LLM maps the text in these sources to nodes, edges, and parameters.

Figure~\ref{fig:plate} shows the plate diagram for the BayesNet of our choice. The network contains three layers of binary nodes representing risk factors $R$, diseases $D$, and symptoms $S$. Risk factors include past medical history and co-morbid conditions, family history, exposures, habits, and travel history. They modulate disease probabilities, while diseases generate observable symptoms. We parameterize both layers using Noisy-OR models \citep{jaakkola1999variational}:
\par\smallskip
\noindent
\begin{minipage}[c]{0.63\linewidth}
\setlength{\abovedisplayskip}{0pt}
\setlength{\belowdisplayskip}{0pt}
\setlength{\abovedisplayshortskip}{0pt}
\setlength{\belowdisplayshortskip}{0pt}
\[
\begin{aligned}
P(D_j{=}1 \mid R) &= 1 - (1 - p_j)\prod_{k\in \mathrm{pa}(j)} (1 - r_{kj})^{R_k}, \\
P(S_i{=}1 \mid D) &= 1 - (1 - \lambda_i)\prod_{j\in \mathrm{pa}(i)} (1 - q_{ji})^{D_j}.
\end{aligned}
\]
\end{minipage}\hfill
\begin{minipage}[c]{0.35\linewidth}
    \centering
    \includegraphics[width=\linewidth]{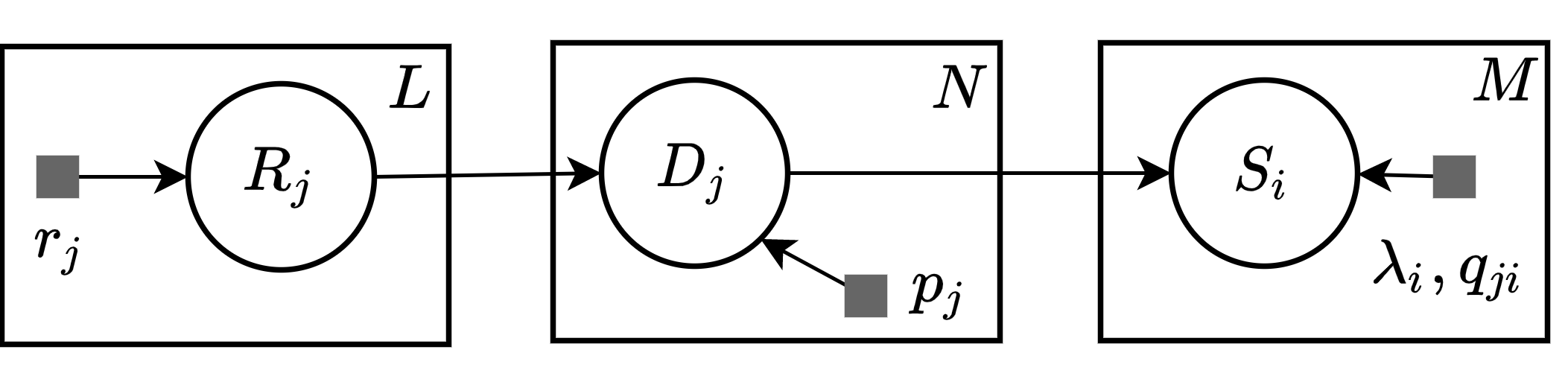}
    \captionof{figure}{Plate diagram for the tripartite Bayesian network.}
    \label{fig:plate}
\end{minipage}
\par\smallskip
Here, $\mathrm{pa}(j)$ and $\mathrm{pa}(i)$ denote the parents of disease $D_j$ and symptom $S_i$, respectively. The parameters $r_{kj}$ and $q_{ji}$ are the strengths of Risk$\rightarrow$Disease and Disease$\rightarrow$Symptom edges, $p_j$ is the baseline probability of $D_j$, and $\lambda_i$ is the symptom leak probability. We estimate $p_j$, $r_{kj}$, and $q_{ji}$ with an LLM and fix every symptom leak to $\lambda_i=0.001$.

At inference time, \bnagent{} accumulates confirmed and denied clinical evidence $ES=(S^+,S^-,R^+,R^-)$ from the dialogue and computes disease posteriors $P(D_j{=}1\mid ES)$. Following \citet{guan2021bayesian}, we use a primary-diagnosis approximation in which one disease is active while the others are inactive:
\begin{equation*}
\begin{split}
P(D_j{=}1 \mid ES) \propto P(D_j{=}1 \mid R)
\prod_{i\in S^+} P(S_i{=}1 \mid D_j)
\prod_{i\in S^-} P(S_i{=}0 \mid D_j).
\end{split}
\end{equation*}
This formulation supports linear-time inference and real-time belief updates during dialogue. Appendix~\ref{app:inference} provides the full derivation. We now describe the three construction stages.

\paragraph{Ontology Building.}
The first stage defines the node set from the clinical variables discussed in the patient notes $\mathcal{P}$. For each note, we prompt an LLM to extract symptoms and five types of risk factors: past medical history, family history, exposures, habits, and travel history. Pooling these mentions produces synonyms and vague terms. In Figure~\ref{fig:bayesnet-construction}, for example, \emph{wheezing} and \emph{wheeze} refer to the same symptom, while \emph{feeling unwell} is too vague to form a useful node.

Passing an entire category to the LLM would create long prompts and require it to compare many unrelated terms. We instead embed the terms using SapBERT \citep{liu-etal-2021-self} and use agglomerative clustering to form small groups of likely matches. We recursively divide clusters larger than 30 terms. The LLM then merges equivalent terms, assigns canonical names, and removes terms that are not useful for diagnosis. Thus, clustering narrows the candidate set, while the LLM makes the final decisions. We repeat this process for $k=3$ rounds. The final symptom category defines $S$, and the other categories form the risk-factor set $R$.

\paragraph{Structure Prediction.}
Once the node set is fixed, we determine which nodes are connected. Co-occurrence in the small note corpus is not reliable evidence of a clinical relation, so we select a medical reference for each disease. For every $D_j\in D$, the LLM generates search queries and fetches\footnote{We use \href{www.tavily.com}{Tavily} API for web searches.} candidate pages from authorative sources such as CDC, NHS and MedlinePlus. Once parsed, the LLM ranks these pages, and we select the highest-ranked page as the source document for $D_j$.

We predict edges separately for each disease-category pair, dividing each category into batches of at most 50 nodes. The LLM retains relations supported by the selected reference and orients them as $R_k\rightarrow D_j$ or $D_j\rightarrow S_i$. In Figure~\ref{fig:bayesnet-construction}, recent travel remains in $R$ because it occurs in the notes, but it has no edge to asthma because the reference does not support that relation. The union of the retained edges forms the final graph structure.

\paragraph{Parameter Estimation.}
With the network structure in place, we estimate the parameters of the Noisy-OR model. Estimating them from $\mathcal{P}$ would be unreliable because the corpus is small and is not intended to provide population-level frequencies. We therefore use the LLM to provide numerical estimates of the parameters. It estimates a baseline probability $p_j$ for every disease, a strength $r_{kj}$ for every retained Risk$\rightarrow$Disease edge, and a strength $q_{ji}$ for every retained Disease$\rightarrow$Symptom edge. We do not estimate strengths for absent edges, and we fix every symptom leak to $\lambda_i=0.001$. This produces a parameterized BayesNet that can be used by \bnagent{} for the inference.

The pipeline can also use expert input when it is available. For example, an expert ontology, graph structure, or parameter set can replace the corresponding stage while the other stages remain unchanged. We study these replacements in our component ablation.
We provide the prompts used for BayesNet construction in Appendix~\ref{app:prompts}.

\section{Experimental Setup}\label{sec:expsetup}

We evaluate \sys{} on principal diagnosis prediction using simulated multi-turn patient interviews. The system conducts a history-taking interview and returns a ranked list of candidate diagnoses. A system performs well when the true primary diagnosis appears near the top of this list.

\paragraph{Patient Simulator and Evaluation.} 
Evaluating medical TOD systems is challenging because several valid questioning paths can lead to the same diagnosis. Existing turn-level protocols, such as action F1 and BLEU \citep{budzianowski2018multiwoz,hosseini2020simple,saley2024meditod}, compare each system turn with a fixed reference dialogue. This is too restrictive for clinical settings. For example, asking about fever before chills can be just as reasonable as asking in the opposite order. These metrics can therefore penalize clinically appropriate behavior and fail to capture the flexibility needed for diagnostic reasoning.

We instead use a dialogue-level evaluation framework. A patient simulator \citep{luo-etal-2024-duetsim,liu-etal-2025-interactive,feng2026doctoragent,PhysioNet-persona-patientsim-1.0.0} interacts with \sys{} over multiple turns. The simulator is powered by Qwen3.8-27B \citep{qwen38} and generates patient responses conditioned on the underlying clinical note. This setup tests whether BayesNet-guided question selection helps \sys{} elicit more diagnostically useful information than unstructured LLM querying. The full simulator prompt is provided in \cref{app:simulator_prompt}. Our standard simulator answers negatively when a queried fact is absent from the clinical note. Because real patients may instead guess, be unsure, or respond ambiguously, we also consider a noisy simulator that randomly gives a plausible diagnosis-consistent answer, reports the fact as unavailable, or expresses uncertainty.

At the end of each simulated interview, \sys{} outputs a ranked list of candidate diagnoses. We compare this list with the true primary diagnosis, encoded using the first three characters of the ICD code. We report Mean Reciprocal Rank (MRR) that gives partial credit when the correct diagnosis is ranked highly and Hit@$k$ \textbf{(}$k=1,3,5$\textbf{)} that measures the proportion of cases where the correct diagnosis appears in the top $k$ predictions. Together, these metrics evaluate diagnostic utility at the dialogue level, without penalizing valid alternative questioning strategies.


\paragraph{Datasets.}
We conduct experiments on DDXPlus and two specialities from MIMIC-IV Notes.
DDXPlus \citep{fansi2022ddxplus} is a dataset of synthetic patient profiles generated from a proprietary medical knowledge base. It covers 49 conditions across multiple specialties. Each profile contains a structured set of symptoms and antecedents, together with a differential diagnosis list. We use 144 cases to construct our BayesNet and 882 cases for evaluation. In addition, we construct \textbf{BayesNet (Oracle)} from the one million patient cases available with DDXPlus. This data-driven network serves as an upper bound on the BayesNet constructed from the small note corpus.

MIMIC-IV Notes\citep{johnson2023mimic} contains de-identified discharge notes from real hospital encounters. As a representative case, we focus on pulmonary diseases, a specialty with clear ICD-coded primary diagnoses, including pneumonia, tuberculosis, and heart failure.
Following \citet{PhysioNet-persona-patientsim-1.0.0}, we use the notes from patients' first visits and map ICD-9-CM codes to ICD-10-CM, retaining only unambiguous one-to-one mappings. We then remove diagnoses with nonspecific qualifiers (e.g., ``unspecified,'' ``unidentified,'' or ``defined elsewhere''). Finally, we use Gemini-3.8-Flash to only retain cases where the diagnosis ICD-10-CM code strongly agrees with the discharge diagnosis from the notes. The resulting \emph{pulmonary} subset contains 17 diseases at the three-character ICD-10-CM level, with 66 notes for BayesNet construction and 197 cases for evaluation.

Finally, we include a second MIMIC-IV Notes specialty, \emph{abdominal}. We use the dataset released by \citet{PhysioNet-mimic-iv-ext-cdm-1.0} consisting of four common diseases. We use 100 cases for BayesNet construction and 599 cases for evaluation. Dataset statistics are summarized in \cref{tab:datasets}.

\paragraph{Baselines.}
Our goal is to test whether \sys{} improves the diagnostic performance of SLMs ($\leq$8B parameters). We evaluate three models: Nemotron-3-Nano-4B (Nano-4B) \citep{nvidia_nemotron3nano4b_2026}, Qwen3-4B, and Qwen3-8B. We compare four configurations for each SLM model. \textbf{Base} performs history taking and diagnosis only using model's capabilities. \textbf{Base + RAG} augments the model with retrieved medical evidence. \textbf{Base + BayesNet} uses the automatically constructed network for belief updates and question selection. \textbf{Base + RAG + BayesNet} combines both forms of augmentation.
On DDXPlus, we additionally evaluate \textbf{Base + BayesNet (Oracle)} to assess the scope of improvement when a large patient corpus available for BayesNet construction. 

For RAG baselines, we use the set of document fetched during BayesNet construction. We break down the documents into chunks and retrieve the top-10 most similar chunks based on language model generated queries. These chunks are added to the model context for generating the next inquiry or the final diagnosis, without blowing up the context size. We provide further details in \cref{app:implementation}.

We evaluate large models, Qwen3-32B \citep{qwen3technicalreport}, Qwen3.5-122B \citep{qwen3.5} and Gemini-3.8-Flash \citep{gemini38flash2026} under Base and Base + BayesNet configurations.

\section{Results}\label{sec:results}
We organize the evaluation around two research questions: a) how much does BayesNet guidance improve diagnostic performance for \sys{}, and where do the gains come from? and b) how effective and reliable is the automatically constructed BayesNet itself?

\subsection{\sys{} Performance and Sources of Gains}
\begin{table}[t]
\centering
\footnotesize
\setlength{\tabcolsep}{3.5pt}
\caption{Results by models across datasets with standard patient simulator. \textbf{Bold} = the best scores withing group. *-results significantly better than paired baseline ($p<0.05$, McnemarTest for Hit@k, WilcoxonTest for MRR).}
\label{tab:main_results_full}
\begin{tabular}{lcccccccc}
\toprule
\textbf{Model} & \multicolumn{4}{c}{\textbf{DDXPlus}} & \multicolumn{4}{c}{\textbf{MIMIC-IV (Pulmonary)}} \\
\cmidrule(lr){2-5} \cmidrule(lr){6-9}
 & \textbf{MRR} & \textbf{Hit@1} & \textbf{Hit@3} & \textbf{Hit@5} & \textbf{MRR} & \textbf{Hit@1} & \textbf{Hit@3} & \textbf{Hit@5} \\
\midrule
Nano~4B & 0.27 & 0.21 & 0.32 & 0.36 & 0.31 & 0.17 & 0.44 & 0.56 \\
\quad + BayesNet & \textbf{0.31}$^{*}$ & \textbf{0.23} & \textbf{0.38}$^{*}$ & \textbf{0.44}$^{*}$ & \textbf{0.37} & \textbf{0.22} & \textbf{0.49} & \textbf{0.61} \\
\addlinespace[1pt]
Nano~4B + RAG & 0.26 & 0.20 & 0.31 & 0.36 & 0.32 & 0.20 & 0.43 & 0.53 \\
\quad + BayesNet & \textbf{0.36*} & \textbf{0.28*} & \textbf{0.43*} & \textbf{0.49*} & \textbf{0.41*} & \textbf{0.27} & \textbf{0.50} & \textbf{0.64*} \\
\midrule
Qwen3~4B & 0.45 & 0.37 & 0.53 & 0.58 & 0.46 & 0.36 & 0.55 & 0.61 \\
\quad + BayesNet & \textbf{0.55}$^{*}$ & \textbf{0.47}$^{*}$ & \textbf{0.63}$^{*}$ & \textbf{0.68}$^{*}$ & \textbf{0.55}$^{*}$ & \textbf{0.45}$^{*}$ & \textbf{0.63}$^{*}$ & \textbf{0.71}$^{*}$ \\
\addlinespace[1pt]
Qwen3~4B + RAG & 0.49 & 0.41 & 0.56 & 0.59 & 0.48 & 0.36 & 0.59 & 0.69 \\
\quad + BayesNet & \textbf{0.56}$^{*}$ & \textbf{0.47}$^{*}$ & \textbf{0.64}$^{*}$ & \textbf{0.69}$^{*}$ & \textbf{0.56}$^{*}$ & \textbf{0.45}$^{*}$ & \textbf{0.65} & \textbf{0.71} \\
\midrule
Qwen3~8B & 0.47 & 0.37 & 0.57 & 0.63 & 0.49 & 0.38 & 0.59 & 0.68 \\
\quad + BayesNet & \textbf{0.56}$^{*}$ & \textbf{0.48}$^{*}$ & \textbf{0.63}$^{*}$ & \textbf{0.67}$^{*}$ & \textbf{0.60}$^{*}$ & \textbf{0.49}$^{*}$ & \textbf{0.67}$^{*}$ & \textbf{0.77}$^{*}$ \\
\addlinespace[1pt]
Qwen3~8B + RAG & 0.49 & 0.41 & 0.55 & 0.60 & \textbf{0.57} & \textbf{0.47} & \textbf{0.68} & 0.71 \\
\quad + BayesNet & \textbf{0.57}$^{*}$ & \textbf{0.49}$^{*}$ & \textbf{0.65}$^{*}$ & \textbf{0.71}$^{*}$ & \textbf{0.57} & 0.46 & 0.66 & \textbf{0.74} \\
\midrule
\midrule
Qwen3~32B & \textbf{0.61} & 0.52 & \textbf{0.70} & \textbf{0.75} & \textbf{0.58} & 0.46 & \textbf{0.69} & \textbf{0.80} \\
\quad + BayesNet & \textbf{0.61} & \textbf{0.53} & 0.69 & 0.74 & \textbf{0.58} & \textbf{0.48} & 0.67 & 0.76 \\
\addlinespace[1pt]
Qwen3~32B + RAG & 0.51 & 0.44 & 0.58 & 0.63 & 0.56 & 0.41 & \textbf{0.71} & \textbf{0.78} \\
\quad + BayesNet & \textbf{0.60}$^{*}$ & \textbf{0.51}$^{*}$ & \textbf{0.69}$^{*}$ & \textbf{0.73}$^{*}$ & \textbf{0.62}$^{*}$ & \textbf{0.53}$^{*}$ & 0.70 & 0.77 \\
\midrule
Qwen3.6~122B & \textbf{0.75} & \textbf{0.66} & \textbf{0.84} & \textbf{0.88} & 0.69 & 0.58 & 0.80 & \textbf{0.87} \\
\quad + BayesNet & \textbf{0.75} & \textbf{0.66} & \textbf{0.84} & 0.87 & \textbf{0.71} & \textbf{0.60} & \textbf{0.82} & \textbf{0.87} \\
\midrule
Gemini~3.8~Flash & \textbf{0.78} & \textbf{0.70} & \textbf{0.85} & \textbf{0.89} & \textbf{0.74} & 0.63 & \textbf{0.84} & \textbf{0.92}$^{*}$ \\
\quad + BayesNet & 0.76 & 0.67 & \textbf{0.85} & \textbf{0.89} & \textbf{0.74} & \textbf{0.64} & 0.81 & 0.88 \\
\bottomrule
\end{tabular}
\end{table}

\paragraph{\sys{} improves performance for SLMs.}
\Cref{tab:main_results_full} shows that \sys{} variants improve performance across all metrics for SLMs on DDXPlus and MIMIC-IV (Pulmonary) datasets. MRR rises by 0.09--0.11 for Qwen3~4B and Qwen3~8B, and by 0.04--0.06 for Nano~4B. The consistence gains across two datasets showcase benefit of \sys{} across various clinical settings.
BayesNet guidance also complements RAG. Relative to the RAG baseline, \sys{} improves MRR in seven of eight model--dataset comparisons and ties in the remaining one. Notably, when paired with larger models, Qwen3~32B,  Qwen3.5~122B and Gemini-3.8-Flash, \sys{} shows little improvement. We conduct a detailed analysis in \cref{app:bn_coverage} to understand why the larger models benefit less from BayesNet guidance. The results suggest that larger models already perform much of the required diagnostic reasoning implicitly, leaving limited margin for explicit BayesNet guidance. We also evaluate \sys{} on the MIMIC-IV (Abdominal) dataset and observe that \sys{} improves performance across SLMs. Results are reported in \cref{tab:abdominal_results}. 

\Cref{app:qualitative} presents qualitative examples of \sys{} in action. Given the patient's chief complaint, the base model overlooks the correct diagnosis and pursues less relevant lines of questioning. Consequently, the final diagnosis is incorrect.
In contrast, the \sys{}-augmented model incorporates the correct diagnosis into its diagnostic belief state early in the interview. This guides more systematic questioning for confirming/ruling out the diagnostic hypotheses, leading to a accurate final diagnosis. 

\paragraph{\sys{} remains robust under noisy evidence.}
While earlier simulations assume clean patient responses, real-world clinical encounters are often incomplete or ambiguous. We thus evaluate \sys{} using a noisy patient simulator that introduces guessing, uncertainty, and uninformative responses. Although evidence noise lowers absolute performance overall, \cref{tab:noisy_results_full} demonstrates consistent relative gains. \sys{} improves MRR across all six SLM comparisons by 0.02--0.07, alongside gains or ties across Hit@$k$ metrics. Results for larger models remain mixed. These findings confirm that the performance gains for SLMs are sustained under noisy conditions.

\paragraph{Both question steering and final diagnosis contribute to gains.}
To isolate whether improvements stem from better evidence gathering during the interview or from integrating the final posterior at diagnosis, we ablate \sys{} incrementally. As shown in \cref{tab:steering_full}, adding BayesNet guidance solely for question steering (\polagent{}) improves MRR across all four SLM--dataset settings by 0.01--0.06.
Additionally providing the BayesNet posterior at diagnosis time (\diagagent{}) yields a further MRR increase of 0.04--0.08 across models. Across all configurations, the complete pipeline achieves the highest performance on every metric, raising overall MRR by 0.09--0.11 over the Base model and confirming that both question steering and posterior-guided diagnosis contribute to the final gain.
We report futher studies regarding comorbidity status and alternative question-selection strategies in \cref{app:comorbidity_results,app:structure_baselines}.

\begin{table*}[]
\centering
\footnotesize
\setlength{\tabcolsep}{3.2pt}
\caption{Effect of BayesNet-guided question steering and diagnosis. \textbf{Bold} marks the best score within each model and dataset.}
\label{tab:steering_full}
\begin{tabular}{llcccccccc}
\toprule
\textbf{Model} & \multicolumn{4}{c}{\textbf{DDXPlus} ($n=882$)} & \multicolumn{4}{c}{\textbf{MIMIC-IV (Pulmonary)} ($n=197$)} \\
\cmidrule(lr){2-5} \cmidrule(lr){6-9}
 & \textbf{MRR} & \textbf{Hit@1} & \textbf{Hit@3} & \textbf{Hit@5} & \textbf{MRR} & \textbf{Hit@1} & \textbf{Hit@3} & \textbf{Hit@5} \\
\midrule
Qwen3~4B & 0.45 & 0.37 & 0.53 & 0.58 & 0.46 & 0.36 & 0.55 & 0.61 \\
\quad + BayesNet Steering & 0.51 & 0.42 & 0.58 & 0.63 & 0.50 & 0.37 & 0.60 & 0.70 \\
\quad + BayesNet Diagnosis & \textbf{0.55} & \textbf{0.47} & \textbf{0.63} & \textbf{0.68} & \textbf{0.55} & \textbf{0.45} & \textbf{0.63} & \textbf{0.71} \\
\midrule
Qwen3~8B & 0.47 & 0.37 & 0.57 & 0.63 & 0.49 & 0.38 & 0.59 & 0.68 \\
\quad + BayesNet Steering & 0.48 & 0.38 & 0.59 & 0.65 & 0.53 & 0.43 & 0.62 & 0.73 \\
\quad + BayesNet Diagnosis & \textbf{0.56} & \textbf{0.48} & \textbf{0.63} & \textbf{0.67} & \textbf{0.60} & \textbf{0.49} & \textbf{0.67} & \textbf{0.77} \\
\bottomrule
\end{tabular}
\end{table*}

\subsection{Evaluation of the Constructed BayesNet}

The gains from BayesNet guidance raise a further question: how well does the automatically constructed network capture the diagnostic relationships that \sys{} relies on? We first assess the network's diagnostic performance and posterior confidence independently of the language models, then use oracle comparisons to examine the quality of its parameters and locate the remaining limitations in construction.

\begin{table}[t]
    \centering
    \caption{Diagnostic performance from direct BayesNet inference on test-set.}
    \label{tab:bayesnet-direct-state-results}
    \begin{tabular}{llrrrrr}
        \toprule
        Dataset & Method & $N$ & MRR & Hit@1 & Hit@3 & Hit@5 \\
        \midrule
        DDXPlus
            & Uniform random & 882 & 0.09 & 0.02 & 0.06 & 0.10 \\
        DDXPlus
            & BayesNet (Gemini-3.8-Flash) & 882
            & \textbf{0.53} & \textbf{0.38} & \textbf{0.64} & \textbf{0.73} \\
        \midrule
        MIMIC-Pulm
            & Uniform random & 197 & 0.19 & 0.06 & 0.17 & 0.28 \\
        MIMIC-Pulm
            & BayesNet (Gemini-3.8-Flash) & 197
            & \textbf{0.45} & \textbf{0.27} & \textbf{0.52} & \textbf{0.70} \\
        \bottomrule
    \end{tabular}
\end{table}

\paragraph{The constructed network provides useful diagnostic signal.}
Given test-set evidence, we rank diseases by their BayesNet posterior probabilities and compare against uniform random ranking over the same disease set. \Cref{tab:bayesnet-direct-state-results} shows the results. On DDXPlus, the BayesNet achieves an MRR of 0.53 and Hit@5 of 0.73, compared with 0.09 and 0.10 for random ranking. On MIMIC-Pulm, the corresponding scores are 0.45 and 0.70, compared with 0.19 and 0.28. The network therefore captures diagnostic relationships that are informative even without the language-model agents.

\begin{table}[t]
    \centering
    \caption{Test-set BayesNet calibration before and after temperature scaling, fitted on training splits ($T=4.99$ for DDXPlus; $T=6.85$ for MIMIC-Pulm). Brier@k measures cumulative top-$k$ posterior mass (lower is better). Scaling improves calibration on both datasets, especially at top-1.}
    \label{tab:bayesnet-temperature-calibration}
    \begin{tabular}{llrrr}
        \toprule
        Dataset & Calibration & Brier@1 & Brier@3 & Brier@5 \\
        \midrule
        DDXPlus
            & Raw BayesNet
            & 0.39 & 0.32 & 0.26 \\
        DDXPlus
            & Temperature-scaled
            & \textbf{0.22} & \textbf{0.22} & \textbf{0.19} \\
        \midrule
        MIMIC-Pulm
            & Raw BayesNet
            & 0.44 & 0.44 & 0.29 \\
        MIMIC-Pulm
            & Temperature-scaled
            & \textbf{0.18} & \textbf{0.24} & \textbf{0.20} \\
        \bottomrule
    \end{tabular}
\end{table}

\paragraph{Posterior Calibration.}
For these diagnostic rankings to support probabilistic decision-making, the confidence assigned to them also matters. To evaluate this confidence, we conduct calibration study using Brier@$k$, comparing the total probability assigned to the top-$k$ diagnoses with whether the correct diagnosis appears in that set. The raw posteriors tend to be overconfident, with Brier@1 scores of 0.39 on DDXPlus and 0.44 on MIMIC-Pulm (\cref{tab:bayesnet-temperature-calibration}). Importantly, fitting a single temperature parameter on each dataset's training split reduces test-set Brier@1 to 0.22 and 0.18, respectively, with improvements also in Brier@3 and Brier@5. The network's diagnostic signal thus comes with confidence estimates that can be improved through a simple calibration step.

\paragraph{Oracle comparisons support the disease--symptom estimates.}
These predictive results establish the network's utility, but leave open how closely its underlying probabilities reflect empirical diagnostic relationships. We compare its parameters with those of the oracle BayesNet constructed from one million DDXPlus cases (\cref{sec:expsetup}). For each parameter, we compute the KL divergence between the oracle and estimated probabilities, then average within each parameter category; lower values indicate closer agreement. Two sampling baselines draw probabilities uniformly at random or from a fitted Gaussian distribution, with results reported over 100 seeded samples. For disease--symptom probabilities, the constructed BayesNet achieves an average KL of 0.38, compared with 0.50 for uniform sampling and 0.41 for Gaussian sampling. This agreement supports the quality of the disease--symptom estimates. However, disease priors and risk factor--disease probabilities are not directly comparable across the two networks: the constructed network estimates them for the general population, including healthy individuals, whereas the oracle estimates them for a population consisting only of patients with a disease. Their KL values, and consequently the overall average, cannot be interpreted directly as estimation accuracy.

\begin{table}[]
\centering
\small
\setlength{\tabcolsep}{4pt}
\caption{BayesNet component ablations on DDXPlus.  \textbf{Bold} = the best score within each model. $^{*}$ = statistically significant difference from the base ($p<0.05$).}
\label{tab:bn_ablation_full}
\begin{tabular}{llcccc}
\toprule
\textbf{Model} & \textbf{Configuration} & \textbf{MRR} & \textbf{Hit@1} & \textbf{Hit@3} & \textbf{Hit@5} \\
\midrule
Qwen3~4B & Base & 0.45 & 0.37 & 0.53 & 0.58 \\
 & \quad + Expert Ontology & 0.54$^{*}$ & 0.47$^{*}$ & 0.60$^{*}$ & 0.66$^{*}$ \\
 & \quad + Expert Structure & 0.67$^{*}$ & 0.59$^{*}$ & 0.74$^{*}$ & 0.79$^{*}$ \\
 & \quad + Expert Parameters & \textbf{0.70}$^{*}$ & \textbf{0.62}$^{*}$ & \textbf{0.78}$^{*}$ & \textbf{0.82}$^{*}$ \\
\midrule
Qwen3~8B & Base & 0.47 & 0.37 & 0.57 & 0.63 \\
 & \quad + Expert Ontology & 0.54$^{*}$ & 0.47$^{*}$ & 0.61$^{*}$ & 0.66$^{*}$ \\
 & \quad + Expert Structure & 0.66$^{*}$ & 0.57$^{*}$ & 0.74$^{*}$ & 0.78$^{*}$ \\
 & \quad + Expert Parameters & \textbf{0.71}$^{*}$ & \textbf{0.63}$^{*}$ & \textbf{0.78}$^{*}$ & \textbf{0.82}$^{*}$ \\
\bottomrule
\end{tabular}
\end{table}

\paragraph{Structure refinement offers the largest gain in the oracle ablation.}
Parameter agreement alone cannot establish which construction errors most affect downstream performance. To assess their impact, we progressively replace the synthesized ontology, structure, and parameters with their oracle counterparts while holding the downstream model fixed (\cref{tab:bn_ablation_full}). Introducing the oracle ontology improves every metric for both Qwen models. The largest subsequent MRR increase comes from replacing the automatically predicted structure: 0.54 to 0.67 for Qwen3~4B and 0.54 to 0.66 for Qwen3~8B. Oracle parameters provide smaller additional increases to 0.70 and 0.71, respectively. Within this cumulative replacement sequence, structure refinement therefore offers the largest gain, making it a promising target for improving automatic construction.

The constructed BayesNet thus supplies useful diagnostic guidance despite imperfections in its construction: its rankings are informative, temperature scaling improves its posterior confidence, and its disease--symptom estimates agree more closely with the oracle than the sampling baselines do. The oracle ablation points to graph structure as a promising route to further downstream gains.

\section{Conclusion}
We propose \sys{}, the first TOD system for medical history taking that integrates SLM's conversational capabilities with interpretable probabilistic reasoning of BayesNets. \sys{} leverages a dedicated \bnagent{} that maintains running diagnostic beliefs over candidate diagnoses and a \polagent{} selects clinically informative questions, thus combining System 1 and System 2 reasoning in a unified framework. To avoid costly curation of BayesNets, \sys{} proposes an automated pipeline that sythesizes BayesNet ontology, structure, and parameters based on given set of diseases and small corpus of patient profiles. The pipeline is flexible, allowing expert intervention at any stage, for further improvements.

We validate \sys{} on two datasets: DDXPlus and MIMIC-IV covering different clinical specialties. Results show that \sys{} consistently improves performance of the paired base models across the metrics, with larger gains for smaller models. Ablation studies confirm importance of each pipeline component, with graph structure learning being the most impactful. BayesNet guidance also leads to more clinically relevant questions and more accurate diagnoses in qualitative analysis. Together, these results highlight the benefits of combining structured Bayesian inference with conversational LLMs, and point toward a more transparent, reliable, and clinically grounded paradigm for medical dialogue systems.


\section*{Acknowledgments}
Mausam is supported by research grants from Google, IBM AI Horizon Networks (AIHN), Wipro, and the Nick McKeown Chair Professorship. Vishal is supported from a Google Fellowship, NeuRIPS travel grant, and the IITD CSE department travel grant. We thank Microsoft AFMR program and Google Cloud Gemini Credits grant for access to LLM API calls. We thank the IIT Delhi HPC facility for its computational resources. This work was partially supported by the CSE Research Acceleration Fund of IIT Delhi

\section*{Limitations}
\label{app:limitations}

While \sys{} demonstrates strong performance across our evaluations, it has several limitations that warrant discussion. \sys{} relies on a predefined set of diseases for constructing the Bayesian network. In practice, clinical settings serve highly diverse patient populations where the relevant disease space may shift over time or vary by region. Supporting such diversity would require automatic or on-the-fly extension of the Bayesian network. Although this direction is both interesting and technically feasible, we leave it to future work.

\sys{} adopts a tripartite Bayesian network structure linking symptoms, diseases, and tests. While this structure is well-motivated for medical dialogue, it is only one of many possible graph topologies. Alternative structures, such as those incorporating risk factors, patient history, or treatment outcomes, may better capture the complexity of real clinical reasoning and are worth exploring in future work.

The Bayesian network in \sys{} is constructed using Gemini 3.8 Flash, a relatively lightweight model chosen for efficiency. Using larger, more capable LLMs for this construction step may yield higher-quality conditional probability estimates and richer graph structures, potentially improving downstream dialogue performance.

Our experiments are conducted exclusively in English. Medical dialogue systems must ultimately serve multilingual populations. Wether the probabilistic reasoning encoded in the Bayesian network transfers effectively to other languages or clinical terminologies is an interesting question.

\section*{Impact Statement}\label{app:impact}
\sys{} advances development of AI systems for medical diagnosis, with potential to improve patient outcomes and reduce healthcare costs. By integrating probabilistic reasoning through Bayesian networks, our approach enhances the accuracy and reliability of diagnostic suggestions in simulated clinical dialogues. This could lead to more informed decision-making by clinicians and better patient care.

Further, our method allows clinical practitioners to specify their own diagnostic knowledge as BayesNet components in structured form, which can be used to guide the dialogue and diagnosis. This is an interesting  human-AI collaboration setting in medical settings, where clinicians can inject their expertise into the system to improve performance and trustworthiness.

Finally, \sys{} works well with small language models, which helps democratize access to advanced diagnostic tools in resource-constrained settings, where large models may be impractical. By enabling effective medical dialogue systems with smaller models, our work has the potential to broaden the reach of AI-driven healthcare solutions and reduce disparities in access to quality care.

\bibliographystyle{abbrvnat}
\bibliography{references}

\appendix
\section{Exact Bayesian Inference using Noisy-OR}\label{app:inference}
Following \citep{guan2021bayesian, cms2024icd10guidelines}, we adopt a primary diagnosis model, i.e., chiefly one disease is responsible for the patient's condition.  Given a clinical evidence state $ES = \{\mathcal{S^+}, \mathcal{S^-}, \mathcal{R^+},\mathcal{R^-}\}$  our goal is to compute $P(D_j=1|ES)$ for each disease $D_j$. We start with marginalizing over all $D_{j^-}$ combinations.
\begin{align*}
    P(D_j=1 \mid \mathcal{S^+}, \mathcal{S^-}, \mathcal{R^+}, \mathcal{R^-}) &= \sum_{\forall D_{j^-}}{P(D_j=1, D_{j^-}|\mathcal{S^+}, \mathcal{S^-}, \mathcal{R^+}, \mathcal{R^-})} \\
    & = P(D_j=1, D_{j^-}=0|\mathcal{S^+}, \mathcal{S^-}, \mathcal{R^+}, \mathcal{R^-})
\end{align*}
Now, posterior is computed by applying Bayes' Rule.
\begin{align*}
    & P(D_j=1 \mid \mathcal{S^+}, \mathcal{S^-}, \mathcal{R^+}, \mathcal{R^-}) \\
    & = \frac{P(\mathcal{S^+}\mathcal{S^-}|D_j=1, D_{j^-}=0) P(D_j=1, D_{j^-}=0, \mathcal{R^+}, \mathcal{R^-})}{P(\mathcal{S^+}, \mathcal{S^-})}
\end{align*}
Note that the term $P(\mathcal{S^+}, \mathcal{S^-})$ is common for all diseases $D_j$. Applying conditional independence of symptoms given disease, we compute the posterior using the following equation.

\begin{align*}
P(D_j = 1 \mid \mathcal{S}^+, \mathcal{S}^-, \mathcal{R^+}, \mathcal{R^-}) \propto 
\; P(D_j = 1, D_{j^-} = 0 \mid \mathcal{R^+}, \mathcal{R^-}) & \nonumber \\
\times \prod_{i \in \mathcal{S}^+} P(S_i = 1 \mid D_j = 1, D_{j^-} = 0) & \nonumber \\
\times \prod_{i \in \mathcal{S}^-} P(S_i = 0 \mid D_j = 1, D_{j^-} = 0) & \nonumber
\end{align*}

where $P(D_j = 1, D_{j^-} = 0 \mid \mathcal{R^+}, \mathcal{R^-}) = P(D_j=1 \mid \mathcal{R^+}, \mathcal{R^-}) \prod_{\forall j' \neq j}{(1 - P(D_{j'} = 1 \mid \mathcal{R^+}, \mathcal{R^-}))}$ are disease probabilities give risk factor combinations. We again apply Noisy-OR model and compute
\begin{align*}
    P(D_j=1 \mid \mathcal{R^+}, \mathcal{R^-}) = 1 - (1-p_j) * \prod_{k\in \text{parent}(D_j))} (1-r_{kj})^{R_k}
\end{align*}
where $p_j$ is prior probability of disease $D_j$ and $r_{kj}$ is conditional likelihood of $D_j$ given risk factor $R_k$

\section{Additional Results}\label{sec:additional_results}

\subsection{Noisy Simulator Results}

\begin{table}[t]
\centering
\footnotesize
\setlength{\tabcolsep}{3.5pt}
\caption{Results by models across datasets with noisy patient simulator. \textbf{Bold} = the best scores withing group. *-results significantly better than paired baseline ($p<0.05$, McnemarTest for Hit@k, WilcoxonTest for MRR).}
\label{tab:noisy_results_full}
\begin{tabular}{lcccccccc}
\toprule
\textbf{Model} & \multicolumn{4}{c}{\textbf{DDXPlus}} & \multicolumn{4}{c}{\textbf{MIMIC-IV (Pulmonary)}} \\
\cmidrule(lr){2-5} \cmidrule(lr){6-9}
 & \textbf{MRR} & \textbf{Hit@1} & \textbf{Hit@3} & \textbf{Hit@5} & \textbf{MRR} & \textbf{Hit@1} & \textbf{Hit@3} & \textbf{Hit@5} \\
\midrule
Nano~4B & 0.24 & 0.18 & 0.29 & 0.34 & 0.28 & 0.16 & 0.36 & 0.49 \\
\quad + BayesNet & \textbf{0.29}$^{*}$ & \textbf{0.21} & \textbf{0.36}$^{*}$ & \textbf{0.43}$^{*}$ & \textbf{0.35}$^{*}$ & \textbf{0.22} & \textbf{0.42} & \textbf{0.59}$^{*}$ \\
\midrule
Qwen3~4B & 0.41 & 0.34 & 0.47 & 0.53 & 0.46 & 0.36 & 0.53 & 0.62 \\
\quad + BayesNet & \textbf{0.43} & \textbf{0.35} & \textbf{0.48} & \textbf{0.56} & \textbf{0.52} & \textbf{0.41} & \textbf{0.62}$^{*}$ & \textbf{0.69} \\
\midrule
Qwen3~8B & 0.44 & 0.34 & 0.52 & 0.59 & 0.49 & 0.37 & 0.58 & \textbf{0.69} \\
\quad + BayesNet & \textbf{0.48}$^{*}$ & \textbf{0.39}$^{*}$ & \textbf{0.56}$^{*}$ & \textbf{0.63} & \textbf{0.52} & \textbf{0.42} & \textbf{0.63} & \textbf{0.69} \\
\midrule
Qwen3~32B & 0.53 & 0.44 & 0.61 & 0.67 & \textbf{0.61} & \textbf{0.50} & \textbf{0.72} & \textbf{0.78} \\
\quad + BayesNet & \textbf{0.55} & \textbf{0.46} & \textbf{0.62} & \textbf{0.68} & 0.59 & 0.48 & 0.68 & \textbf{0.78} \\
\midrule
Qwen3.6~122B & \textbf{0.66} & \textbf{0.57} & \textbf{0.76} & \textbf{0.82} & \textbf{0.72} & \textbf{0.60} & \textbf{0.83} & \textbf{0.89} \\
\quad + BayesNet & \textbf{0.66} & 0.56 & \textbf{0.76} & 0.80 & 0.69 & 0.57 & 0.81 & 0.88 \\
\midrule
Gemini~3.8~Flash & \textbf{0.71} & \textbf{0.62} & \textbf{0.80} & \textbf{0.85} & \textbf{0.73} & \textbf{0.62} & \textbf{0.85} & \textbf{0.91} \\
\quad + BayesNet & 0.69 & 0.59 & 0.78 & 0.84 & 0.69 & 0.57 & 0.81 & 0.88 \\
\bottomrule
\end{tabular}
\end{table}

\subsection{Results by Comorbidity Status}\label{app:comorbidity_results}
\begin{table}[]
\centering
\footnotesize
\setlength{\tabcolsep}{3.5pt}
\caption{MRR on DDXPlus, stratified by comorbidity status. \textbf{Bold} marks the higher score within each paired baseline--BayesNet comparison.}
\label{tab:comorbidity}
\begin{tabular}{lccc}
\toprule
\textbf{Model} & \multicolumn{3}{c}{\textbf{DDXPlus}} \\
\cmidrule(lr){2-4}
 & \shortstack{\textbf{All} \\ {\scriptsize($N$=882)}} & \shortstack{\textbf{Comorbid} \\ {\scriptsize($N$=387)}} & \shortstack{\textbf{No-comorbid} \\ {\scriptsize($N$=495)}} \\
\midrule
Nano~4B & 0.27 & 0.29 & 0.24 \\
\quad + BayesNet & \textbf{0.31} & \textbf{0.34} & \textbf{0.28} \\
\midrule
Qwen3~4B & 0.45 & 0.48 & 0.43 \\
\quad + BayesNet & \textbf{0.55} & \textbf{0.59} & \textbf{0.53} \\
\midrule
Qwen3~8B & 0.47 & 0.49 & 0.45 \\
\quad + BayesNet & \textbf{0.56} & \textbf{0.59} & \textbf{0.53} \\
\midrule
Qwen3~32B & \textbf{0.61} & 0.59 & \textbf{0.63} \\
\quad + BayesNet & \textbf{0.61} & \textbf{0.62} & 0.60 \\
\midrule
Gemini~3.8~Flash & \textbf{0.78} & 0.74 & \textbf{0.81} \\
\quad + BayesNet & 0.76 & \textbf{0.75} & 0.77 \\
\bottomrule
\end{tabular}

\end{table}

\paragraph{\sys{} benefits SLMs with and without comorbidities.}
\Cref{tab:comorbidity} stratifies the DDXPlus results by whether a patient has a comorbid condition. BayesNet guidance improves MRR for all three SLMs in both groups: by 0.05--0.11 on comorbid cases and by 0.04--0.10 on cases without comorbidities. Thus, the overall improvements are driven by both the patients with and without comorbidities. The larger models again exhibit limited headroom: their small gains on comorbid cases do not transfer to the no-comorbidity group, consistent with the scale-dependent trend in the main results.

\subsection{Results on the MIMIC-IV (Abdominal) Dataset}\label{app:abdominal_results}
\Cref{tab:abdominal_results} showcases performance of \sys{} on MIMIC-IV (Abdominal) dataset. Since abdominal dataset only has four target pathologies, we adopt top-1 diagnostic accuracy as the evaluation metric. The results reported in \Cref{tab:abdominal_results} show trend similar to the DDXPlus and MIMIC-IV (Pulmonary) datasets. Notably, \sys{} improves performance of Base models by and RAG settings for SLMs.

\begin{table}[]
\centering
\small
\setlength{\tabcolsep}{5pt}
\caption{Results MIMIC-IV (Abdominal) dataset. \textbf{Bold} = the best scores withing group. .*-results significantly better than base model ($p<0.05$ Mcnemar Test).}
\label{tab:abdominal_results}
\begin{tabular}{lc}
\toprule
\textbf{Model} & \textbf{Hit\@1} \\
\midrule
Nano~4B & 0.43 \\
Nano~4B + BayesNet & \textbf{0.48}$^{*}$ \\
\addlinespace[2pt]
Nano~4B + RAG & 0.44 \\
Nano~4B + RAG + BayesNet & \textbf{0.48} \\
\midrule
Qwen3~4B & 0.58 \\
Qwen3~4B + BayesNet & \textbf{0.60} \\
\addlinespace[2pt]
Qwen3~4B + RAG & 0.55 \\
Qwen3~4B + RAG + BayesNet & \textbf{0.63}$^{*}$ \\
\midrule
Qwen3~8B & 0.52 \\
Qwen3~8B + BayesNet & \textbf{0.57}$^{*}$ \\
\addlinespace[2pt]
Qwen3~8B + RAG & 0.57 \\
Qwen3~8B + RAG + BayesNet & \textbf{0.61} \\
\midrule
Qwen3~32B & \textbf{0.75} \\
Qwen3~32B + BayesNet & \textbf{0.75} \\
\addlinespace[2pt]
Qwen3~32B + RAG & \textbf{0.79} \\
Qwen3~32B + RAG + BayesNet & 0.75$^{*}$ \\
\midrule
Qwen3.6~122B & 0.76 \\
Qwen3.6~122B + BayesNet & \textbf{0.77} \\
\midrule
Gemini~3.8~Flash & \textbf{0.79} \\
Gemini~3.8~Flash + BayesNet & 0.75$^{*}$ \\
\bottomrule
\end{tabular}
\end{table}

\paragraph{Guidance improves smaller models in the clean setting.}
BayesNet improves top-1 accuracy for Nano~4B, Qwen3~4B, and Qwen3~8B both with and without RAG, with gains of 0.02--0.08. The largest gain is for Qwen3~4B with RAG, from 0.55 to 0.63. Results for larger models are mixed: Qwen3~32B ties its baseline at 0.75 but drops from 0.79 to 0.75 with RAG, Qwen3.6~122B improves from 0.76 to 0.77, and Gemini~3.8~Flash drops from 0.79 to 0.75.

\paragraph{Benefits are less consistent in the noisy setting.}
Guidance improves Nano~4B from 0.45 to 0.47 and Qwen3.6~122B from 0.80 to 0.81, but reduces accuracy for the other four models. The largest decrease is for Qwen3~8B, from 0.64 to 0.58. Thus, the gains observed for smaller models in the clean abdominal setting do not consistently carry over to the noisy patient simulator.

\subsection{Alternative Question-Selection Strategies}\label{app:structure_baselines}
We compare \sys{} with two alternative strategies. The Naive Bayes variant reuses the disease--symptom edges and parameter estimates constructed by \sys{}, but adopts a simpler probabilistic structure. Information Gain instead greedily selects the question with the largest expected one-step reduction in diagnostic uncertainty.

\begin{table*}[]
\centering
\footnotesize
\setlength{\tabcolsep}{3.2pt}
\caption{Comparison of question-selection structures on DDXPlus and MIMIC-IV (Pulmonary). \textbf{Bold} marks the best score within each model and dataset.}
\label{tab:structure_baselines_full}
\begin{tabular}{llcccccccc}
\toprule
\textbf{Model} & \textbf{Method} & \multicolumn{4}{c}{\textbf{DDXPlus} ($n=882$)} & \multicolumn{4}{c}{\textbf{MIMIC-IV (Pulmonary)} ($n=197$)} \\
\cmidrule(lr){3-6} \cmidrule(lr){7-10}
 &  & \textbf{MRR} & \textbf{Hit@1} & \textbf{Hit@3} & \textbf{Hit@5} & \textbf{MRR} & \textbf{Hit@1} & \textbf{Hit@3} & \textbf{Hit@5} \\
\midrule
Nano~4B & Base & 0.27 & 0.21 & 0.32 & 0.36 & 0.31 & 0.17 & 0.44 & 0.56 \\
 & Naive Bayes & 0.29 & 0.22 & 0.36 & 0.42 & 0.33 & 0.21 & 0.45 & 0.54 \\
 & Information Gain & \textbf{0.39} & \textbf{0.30} & \textbf{0.48} & \textbf{0.53} & \textbf{0.37} & \textbf{0.24} & 0.47 & 0.60 \\
 & BayesNet & 0.31 & 0.23 & 0.38 & 0.44 & \textbf{0.37} & 0.22 & \textbf{0.49} & \textbf{0.61} \\
\midrule
Qwen3~4B & Base & 0.45 & 0.37 & 0.53 & 0.58 & 0.46 & 0.36 & 0.55 & 0.61 \\
 & Naive Bayes & 0.51 & 0.43 & 0.59 & 0.63 & 0.50 & 0.39 & 0.58 & 0.68 \\
 & Information Gain & 0.46 & 0.37 & 0.54 & 0.60 & 0.45 & 0.31 & 0.56 & 0.69 \\
 & BayesNet & \textbf{0.55} & \textbf{0.47} & \textbf{0.63} & \textbf{0.68} & \textbf{0.55} & \textbf{0.45} & \textbf{0.63} & \textbf{0.71} \\
\midrule
Qwen3~8B & Base & 0.47 & 0.37 & 0.57 & 0.63 & 0.49 & 0.38 & 0.59 & 0.68 \\
 & Naive Bayes & 0.54 & 0.46 & 0.61 & \textbf{0.67} & 0.52 & 0.44 & 0.58 & 0.66 \\
 & Information Gain & 0.48 & 0.38 & 0.57 & 0.63 & 0.45 & 0.31 & 0.57 & 0.69 \\
 & BayesNet & \textbf{0.56} & \textbf{0.48} & \textbf{0.63} & \textbf{0.67} & \textbf{0.60} & \textbf{0.49} & \textbf{0.67} & \textbf{0.77} \\
\midrule
Qwen3~32B & Base & \textbf{0.61} & 0.52 & \textbf{0.70} & \textbf{0.75} & 0.58 & 0.46 & \textbf{0.69} & \textbf{0.80} \\
 & Naive Bayes & 0.60 & 0.50 & 0.68 & 0.74 & \textbf{0.60} & \textbf{0.51} & 0.66 & 0.76 \\
 & Information Gain & 0.51 & 0.41 & 0.61 & 0.67 & 0.54 & 0.39 & 0.66 & 0.79 \\
 & BayesNet & \textbf{0.61} & \textbf{0.53} & 0.69 & 0.74 & 0.58 & 0.48 & 0.67 & 0.76 \\
\bottomrule
\end{tabular}
\end{table*}

\paragraph{The learned disease--symptom model is useful even in a simpler structure.}
As shown in \cref{tab:structure_baselines_full}, Naive Bayes improves MRR over the Base model in all six SLM--dataset comparisons, by 0.02--0.07. On DDXPlus, its gains for Qwen3~4B and Qwen3~8B are significant across every metric. Because this variant uses the disease--symptom relations and estimates produced by our construction pipeline, its strong performance provides additional evidence that these learned components capture useful clinical dependencies. The complete BayesNet nevertheless achieves higher MRR than Naive Bayes in all six SLM settings, indicating that the full structure provides further value.

\paragraph{Greedy information gain is not aligned with final diagnostic accuracy.}
Information Gain trails the complete BayesNet in six of eight model--dataset comparisons and ties it in one, with Nano~4B on DDXPlus as the sole exception. For Qwen3~4B and Qwen3~8B, it yields little or no improvement over Base and is consistently below the complete BayesNet. This result is expected from its myopic objective: a question that maximally reduces uncertainty at the current turn need not produce the most useful sequence of evidence for the final diagnosis. In contrast, \sys{} allows the policy model to interpret the BayesNet belief jointly with the dialogue history when selecting the next question.

\subsection{Diminishing Gains for Larger Models}\label{app:bn_coverage}

We examine why Bayesian Network (BN) guidance benefits smaller models such as Qwen3 4B more than the stronger Gemini 3.8 Flash. To understand this difference, we compare baselines and corresponding \sys{} predictions on sets of cases stratified by whether the gold diagnosis appears in the final BayesNet differential. Table~\ref{tab:bn_coverage} summarizes the results across DDXPlus cases and MIMIC-IV (Pulmonary) cases.

\begin{table}[]
\centering
\small
\caption{MRR with and without BayesNet. Gold coverage denotes the percentage of cases where the final BayesNet differential includes the gold diagnosis. Changes are computed within gold-present and gold-absent groups.}
\label{tab:bn_coverage}
\setlength{\tabcolsep}{3.5pt}
\begin{tabularx}{\linewidth}{@{}ll*{5}{>{\raggedleft\arraybackslash}X}@{}}
\hline
Dataset & Model & \shortstack[r]{Baseline\\MRR} & \shortstack[r]{With\\BayesNet} & \shortstack[r]{Gold\\Coverage (\%)}
& \shortstack[r]{$\Delta$ Gold\\Present} & \shortstack[r]{$\Delta$ Gold\\Absent} \\
\hline
DDXPlus & Qwen3 4B
& 0.45 & 0.55 & 70.98 & +0.21 & -0.17 \\
DDXPlus & Gemini 3.8 Flash
& 0.78 & 0.76 & 77.89 & +0.05 & -0.23 \\
MIMIC-IV (Pulmonary) & Qwen3 4B
& 0.46 & 0.55 & 81.73 & +0.14 & -0.13 \\
MIMIC-IV (Pulmonary) & Gemini 3.8 Flash
& 0.74 & 0.74 & 90.86 & +0.01 & -0.23 \\
\hline
\end{tabularx}
\end{table}

The results suggest that stronger baseline performance leaves less room for useful BayesNet guidance. When the BayesNet includes the correct diagnosis, Qwen3 4B gains 0.21 MRR on DDXPlus and 0.14 on MIMIC-IV (Pulmonary), whereas Gemini gains only 0.05 and 0.01, respectively.
This pattern is consistent with the BayesNet supplying diagnostic candidates that the smaller model would otherwise miss or rank poorly, while providing information that the stronger model can already support on their own.

The reduced benefit does not coincide with lower BayesNet coverage. Gemini has higher gold coverage on both datasets. Nevertheless, excluding the correct diagnosis remains associated with losses, reaching 0.23 MRR for Gemini on each dataset. Thus, as baseline capability increases, the additional value of helpful guidance shrinks while the cost associated with misleading guidance persists. This balance explains the substantial net gains for Qwen3 4B and the absence of comparable gains for Gemini.

\section{Implementation Details}\label{app:implementation}
We run all experiments on up to 8 NVIDIA A100 80GB GPUs. We use vLLM for inference over base language models. Open-weight models, especially SLMs, can produce repetitive outputs or violate the required response format. To reduce these failures, we use vLLM structured generation with temperature set to 0.0, and our prompts explicitly forbid verbatim repetition. If a model violates the format or repeats itself, we resample up to five times with increasing temperature values: 0.1, 0.2, 0.3, 0.5, and 0.7. If no valid response is produced after five attempts, we halt the simulation. 
All our models run in bfloat16 precision without quantization. We run multiple simulations in parallel to reduce evaluation time.

For RAG baselines, we use the same set of document fetched during BayesNet construction. We divide these documents into overlapping chunks of 512 tokens. At each dialogue turn, the language model generates retrieval queries from the conversation so far. We embed the queries and chunks with the SapBERT (\texttt{cambridgeltl/SapBERT-from-PubMedBERT-fulltext}) \citep{liu-etal-2021-self} encoder and retrieve the 10 most similar chunks based on cosine similarity. These chunks are added to the model context for generating the next inquiry or the final diagnosis.

All our simulations take about 3 hours to run per model per dataset. BayesNet synthesis is a one time operation that takes about 4 hour per dataset. The majority of this time is spent on parameter estimation, which requires multiple calls to the LLM for each disease-symptom and risk-factor disease pair.

Note that DDXPlus is released under CC BY 4.0 license while MIMIC-IV is released under a custom license that requires credentialing and data use agreement. We have obtained necessary permissions to use these datasets for research purposes.
\section{Dataset and Bayesnet Statistics}
Table~\ref{tab:datasets} shows the statistics of the datasets used in our experiments. Table~\ref{tab:bn_table} shows the statistics of the Bayesnets used in our experiments, including the number of nodes, edges, and the average Markov blanket size.

\begin{table}[]
\centering
\caption{Dataset statistics for evaluation.}
\label{tab:datasets}
\begin{tabular}{lccc}
\toprule
\textbf{Dataset} & \textbf{Train \# Cases} & \textbf{Test \# Cases} & \textbf{\# Diagnoses} \\
\midrule
DDXPlus        & 144 & 882 & 49 \\
MIMIC-IV Notes (Pulmonary) & 66 & 197 & 17 \\
MIMIC-IV Notes (Abdominal) & 100 & 599 & 4 \\
\bottomrule
\end{tabular}
\end{table}

\begin{table}[]
\centering
\caption{BayesNet Statistics}
\label{tab:bn_table}
\small
\begin{tabularx}{\linewidth}{>{\raggedright\arraybackslash}Xccc}
\toprule
\textbf{Dataset} & \textbf{DDXPlus} & \textbf{MIMIC-IV (Pulmonary)} & \textbf{MIMIC-IV (Abdominal)} \\ \midrule
\#Disease                & 49      & 17     & 4      \\
\#Symptom                & 104     & 47     & 13     \\
\#Risk Factor            & 70      & 47     & 15     \\
\#Disease to Symptom     & 621     & 178    & 28     \\
\#Risk Factor to Disease & 218     & 149    & 19     \\
Construction Cost        & \$16.28 & \$7.94 & \$5.69 \\ \bottomrule
\end{tabularx}
\end{table}
\clearpage
\section{Qualitative Examples}
\label{app:qualitative}

\subsection{PULM: Qwen3-8B, case 16}
\label{app:example:pulm_qwen3_8b_16}
\paragraph{Case Summary.} Patient has worsening shortness of breath for one day, preceded by three days of cough, congestion, low-grade fever, and night sweats. Increasing cough with mild hemoptysis, exertional dyspnea worse than baseline, and chest tightness temporarily relieved by inhalers. Similar episodes occur each winter.

Patient has a medical history of Bronchiectasis, prior right lower lobectomy, and reactive airway disease. Baseline exertional dyspnea limits walking around the house.

\textbf{Gold Diagnosis:} Bronchiectasis

The interviews below compare Qwen3-4B and Qwen3-4B+\sys{} interacting with the simulated patient described above. Both models receive the same chief complaint: shortness of breath. Qwen3-4B begins by asking about the nature and duration of the breathing difficulty and associated symptoms. It subsequently focuses on travel history, immobility, and a family history of blood clots, incorrectly favoring pulmonary embolism. The model neither asks about prior lung disease or surgery nor identifies the gold diagnosis of bronchiectasis.

In contrast, Qwen3-4B+\sys{} begins by asking about chronic cough and smoking history. Although bronchiectasis is absent from the initial Bayesian network belief state, the model quickly considers it based on chronic cough with blood-tinged sputum and the absence of a smoking history. It then systematically gathers evidence to assess competing diagnostic hypotheses. For example, it asks about fever and chills in turn 2 to assess tuberculosis and elicits a history of recurrent infections and lung disease to investigate bronchiectasis. This methodical history-taking leads to the gold diagnosis.

\par\medskip
\noindent\textbf{Chief Complaint:} Doctor, I'm really struggling to catch my breath.\par\medskip

\begingroup
\newbox\caseleft \newbox\caseright
\newbox\caseleftpart \newbox\caserightpart
\newdimen\casewidth \newdimen\caseheight \newdimen\caseremaining
\casewidth=0.487\linewidth
\splittopskip=0pt \splitmaxdepth=0pt \vbadness=10000
\long\def\casecolumn#1{\vbox{\hsize=\dimexpr\casewidth-14pt\relax
\linewidth=\hsize \columnwidth=\hsize
\small\raggedright\setlength{\parindent}{0pt}#1\par}}
\def\casepages{%
  \caseheight=\dimexpr\pagegoal-\pagetotal-40pt\relax
  \ifdim\caseheight>\dimexpr\textheight-40pt\relax
    \caseheight=\dimexpr\textheight-40pt\relax\fi
  \ifdim\caseheight<80pt \newpage
    \caseheight=\dimexpr\textheight-40pt\relax\fi
  \caseremaining=\dimexpr\ht\caseleft+\dp\caseleft\relax
  \ifdim\dimexpr\ht\caseright+\dp\caseright\relax>\caseremaining
    \caseremaining=\dimexpr\ht\caseright+\dp\caseright\relax\fi
  \ifdim\caseremaining<\caseheight \caseheight=\caseremaining\fi
  \setbox\caseleftpart=\vsplit\caseleft to \caseheight
  \setbox\caserightpart=\vsplit\caseright to \caseheight
  \setbox\caseleftpart=\vbox{\unvbox\caseleftpart}
  \setbox\caserightpart=\vbox{\unvbox\caserightpart}
  \noindent\begin{minipage}[t]{\casewidth}
    \begin{tcolorbox}[enhanced,colback=basecolor,colframe=black!45,boxrule=0.4pt,
      before skip=0pt,after skip=0pt,left=6pt,right=6pt,top=5pt,bottom=5pt]
      \vbox to \caseheight{\unvbox\caseleftpart\vfil}
    \end{tcolorbox}
  \end{minipage}\hfill
  \begin{minipage}[t]{\casewidth}
    \begin{tcolorbox}[enhanced,colback=bncolor,colframe=black!45,boxrule=0.4pt,
      before skip=0pt,after skip=0pt,left=6pt,right=6pt,top=5pt,bottom=5pt]
      \vbox to \caseheight{\unvbox\caserightpart\vfil}
    \end{tcolorbox}
  \end{minipage}\par
  \ifvoid\caseleft
    \ifvoid\caseright\else\newpage\casepages\fi
  \else\newpage\casepages\fi}

\setbox\caseleft=\casecolumn{
\textbf{Base (13 doctor turns)}\par\smallskip\hrule\medskip
\textbf{Dr (1):} Can you describe the nature of your breathing difficulty?\par\smallskip
\textbf{Pt:} It’s really tight in my chest, and I feel like I can’t get enough air. Even just getting out of bed or walking to the door makes me winded. I used to be able to walk around my house without stopping, but now I have to stop and catch my breath almost immediately.\par\smallskip
\textbf{Dr (2):} How long have you been experiencing this difficulty breathing, and has it been getting worse over time?\par\smallskip
\textbf{Pt:} It started about three days ago with a cold—cough, congestion, and a little fever. But the breathing really got bad just yesterday. It’s been getting worse since then, not better.\par\smallskip
\textbf{Dr (3):} Have you had any recent travel, or do you have a history of cardiovascular disease or smoking?\par\smallskip
\textbf{Pt:} No, I haven't been traveling anywhere. I don't have any heart problems, and I don't smoke.\par\smallskip
\textbf{Dr (4):} Have you been producing any sputum, and if so, what is its color, consistency, and volume?\par\smallskip
\textbf{Pt:} Yeah, I've been coughing up some stuff. It's not a lot, just a little bit, but it's got some dark blood in it. It's not a huge amount, just a few streaks or drops when I cough.\par\smallskip
\par\smallskip{\color{black!60}\itshape [Doctor turn 5 omitted]}\par\smallskip

\textbf{Dr (6):} Have you been immobile or bedridden for any period recently, or do you have a family history of blood clots or deep vein thrombosis?\par\smallskip
\textbf{Pt:} No, I haven't been stuck in bed. I've been moving around, just slower than usual because of the breathing. As for family history, I don't know of anyone having blood clots or DVT. My family has had some cancers and heart issues, but not clots.\par\smallskip
\par\smallskip{\color{black!60}\itshape [Doctor turns 7--11 omitted]}\par\smallskip

\textbf{Dr (12):} Have you been on any long trips or had prolonged immobility recently that could increase your risk for blood clots?\par\smallskip
\textbf{Pt:} No, I haven't been on any long trips. And I haven't been stuck in one place for a long time. I've been getting up and moving around, just not as much as I usually do because of the breathing.\par\smallskip
\textbf{Dr (13):} Thank you for sharing all these details. I believe we have covered all the necessary information to proceed with the next steps. Let me review the findings and prepare for the next phase.\par\smallskip
\medskip\textbf{Final diagnosis turn}\par\smallskip
\textbf{Reasoning:} The patient presents with acute dyspnea that has progressively worsened over the past three days, accompanied by cough, congestion, low-grade fever, and night sweats. The presence of blood-tinged sputum, along with a history of no recent travel, immobility, or risk factors for pulmonary embolism, suggests that a pulmonary embolism may not be the primary concern. However, the progressive nature of dyspnea, nocturnal sweats, and fever raise a suspicion of an infectious process. The absence of a history of smoking or heart disease makes heart failure less likely, although it cannot be entirely ruled out without further evaluation. The presence of blood-tinged sputum and fever could point towards an infectious etiology, such as acute pharyngitis or tonsillitis, but the severity of respiratory symptoms suggests a more significant condition. The possibility of tuberculosis is noted given the presence of fever, night sweats, and cough, although the patient denies exposure. The presence of chest tightness and respiratory discomfort also raises the concern of acute bronchitis or even pneumonitis. Finally, the acute nature of symptoms, along with the absence of other risk factors, makes pulmonary embolism a possible differential, though the lack of immobility or other risk factors decreases its likelihood.\par\smallskip
\textbf{Differential:}
\begin{enumerate}[leftmargin=*,nosep,label=\arabic*.]
\item Pulmonary embolism
\item Acute bronchitis
\item Acute pharyngitis
\item Respiratory tuberculosis
\item Pneumonitis due to solids and liquids
\end{enumerate}
}
\setbox\caseright=\casecolumn{
\textbf{\sys{} (14 doctor turns)}\par\smallskip\hrule\medskip
\par{\footnotesize\textit{Diagnostic Belief State: Asthma (37.71\%); Heart failure (26.40\%); Pneumonitis due to solids and liquids (13.23\%); Bronchiectasis (8.35\%); Emphysema (6.55\%).}}\par
\textbf{Dr (1):} Have you had a chronic cough or a history of smoking?\par\smallskip
\textbf{Pt:} I do have a cough, yes. It's been bothering me for a few days now, and I've been coughing up a little bit of blood. But no, I don't smoke. I haven't smoked in a long time.\par\smallskip
\par{\footnotesize\textit{Diagnostic Belief State: Bronchiectasis (65.56\%); Pneumonitis due to solids and liquids (21.31\%); Pulmonary edema (6.82\%); Abscess of lung and mediastinum (3.42\%); Respiratory tuberculosis (2.10\%).}}\par
\textbf{Dr (2):} Have you had any fever or chills along with your cough?\par\smallskip
\textbf{Pt:} Yes, I've had a low-grade fever. I've also been breaking out in night sweats, which is pretty unusual for me.\par\smallskip
\par\smallskip{\color{black!60}\itshape [Doctor turns 3--11 omitted]}\par\smallskip

\par{\footnotesize\textit{Diagnostic Belief State: Bronchiectasis (57.42\%); Pneumonitis due to solids and liquids (32.78\%); Pulmonary edema (5.47\%); Asthma (3.94\%); Respiratory tuberculosis (0.23\%).}}\par
\textbf{Dr (12):} Have you had a history of recurrent respiratory infections or a previous diagnosis of lung disease?\par\smallskip
\textbf{Pt:} Yes, I have a lung condition called bronchiectasis. It’s been with me for a long time. I actually had part of my right lung removed when I was a child because of it. I usually get these breathing problems once a year, mostly in the winter, but this time it feels a bit different.\par\smallskip
\par{\footnotesize\textit{Diagnostic Belief State: Bronchiectasis (57.42\%); Pneumonitis due to solids and liquids (32.78\%); Pulmonary edema (5.47\%); Asthma (3.94\%); Respiratory tuberculosis (0.23\%).}}\par
\textbf{Dr (13):} Have you experienced any recent changes in the frequency or severity of your breathing difficulties, or have you had any episodes of chest pain or dizziness?\par\smallskip
\textbf{Pt:} My breathing has definitely gotten worse. I used to be able to walk around my house before I had to stop, but now I get short of breath just getting out of bed or walking to the door. I don't have any chest pain, just that tightness, and I haven't felt dizzy at all.\par\smallskip
\par{\footnotesize\textit{Diagnostic Belief State: Bronchiectasis (63.77\%); Pneumonitis due to solids and liquids (25.48\%); Pulmonary edema (6.07\%); Asthma (4.38\%); Respiratory tuberculosis (0.15\%).}}\par
\textbf{Dr (14):} Thank you for your time. I have gathered enough information to proceed with the next steps in your care.\par\smallskip
\medskip\textbf{Final diagnosis turn}\par\smallskip
\par{\footnotesize\textit{Diagnostic Belief State: Bronchiectasis (63.77\%); Pneumonitis due to solids and liquids (25.48\%); Pulmonary edema (6.07\%); Asthma (4.38\%); Respiratory tuberculosis (0.15\%).}}\par
\textbf{Reasoning:} The patient presents with dyspnea, cough with hemoptysis, low-grade fever, night sweats, chest tightness, and a history of bronchiectasis. These symptoms suggest chronic lung disease complications. The Bayesian Network Hints include Bronchiectasis, Asthma, and Pulmonary edema, which align with the patient's history. Given the patient's known bronchiectasis, this condition is a top priority. The intermittent fever, night sweats, and hemoptysis raise concern for infection, which could include tuberculosis. The wheezing and response to inhalers suggest asthma, though this may be secondary to the bronchiectasis. The patient's denial of leg swelling and weight gain makes pulmonary edema less likely. The history of bronchiectasis and acute exacerbation supports its inclusion in the differential. Asthma is considered given the wheezing and inhaler use. Pulmonary edema is included as a possibility due to the respiratory symptoms, though the absence of peripheral edema makes it less likely. The presence of hemoptysis, fever, and night sweats supports respiratory tuberculosis as a possibility, though the patient denies exposure. Pneumonitis due to solids and liquids is included given the history of recent respiratory infection, though the absence of aspiration risk factors makes it less likely.\par\smallskip
\textbf{Differential:}
\begin{enumerate}[leftmargin=*,nosep,label=\arabic*.]
\item Bronchiectasis
\item Asthma
\item Respiratory tuberculosis
\item Pulmonary edema
\item Pneumonitis due to solids and liquids
\end{enumerate}
}
\casepages
\endgroup

\subsection{DDXPLUS: Qwen3-8B, case 1760}
\label{app:example:ddxplus_qwen3_8b_1760}
\paragraph{Case Summary.} Patient is a 49-year-old woman with vomiting after coughing.
Patient reports exposure to someone with pertussis (whooping cough). She reports being significantly overweight for her height. Vaccinations are up to date.

\textbf{Gold Diagnosis:} Whooping cough

The interviews below compare Qwen3-4B and Qwen3-4B+\sys{} interacting with the simulated patient described above. Both models receive the same chief complaint: vomiting after coughing. Qwen3-4B begins by asking about the duration of the cough and accompanying symptoms. It then explores several diagnostic possibilities, asking about heartburn (GERD), recent chest trauma (spontaneous rib fracture), and stridor (croup). However, it neither considers whooping cough nor asks about exposure to sick individuals.

In contrast, Qwen3-4B+\sys{} identifies whooping cough as a likely diagnosis in its Bayesian network–driven diagnostic belief state. Its subsequent questions focus on gathering evidence to distinguish among competing hypotheses. For example, it immediately asks about exposure to sick individuals, explores dietary changes to assess possible anaphylaxis, and asks about night sweats, fatigue, and weight loss to evaluate other systemic causes. This targeted questioning leads to the correct diagnosis.

\par\medskip
\noindent\textbf{Chief Complaint:} I vomit after coughing.\par\medskip

\begingroup
\newbox\caseleft \newbox\caseright
\newbox\caseleftpart \newbox\caserightpart
\newdimen\casewidth \newdimen\caseheight \newdimen\caseremaining
\casewidth=0.487\linewidth
\splittopskip=0pt \splitmaxdepth=0pt \vbadness=10000
\long\def\casecolumn#1{\vbox{\hsize=\dimexpr\casewidth-14pt\relax
\linewidth=\hsize \columnwidth=\hsize
\small\raggedright\setlength{\parindent}{0pt}#1\par}}
\def\casepages{%
  \caseheight=\dimexpr\pagegoal-\pagetotal-40pt\relax
  \ifdim\caseheight>\dimexpr\textheight-40pt\relax
    \caseheight=\dimexpr\textheight-40pt\relax\fi
  \ifdim\caseheight<80pt \newpage
    \caseheight=\dimexpr\textheight-40pt\relax\fi
  \caseremaining=\dimexpr\ht\caseleft+\dp\caseleft\relax
  \ifdim\dimexpr\ht\caseright+\dp\caseright\relax>\caseremaining
    \caseremaining=\dimexpr\ht\caseright+\dp\caseright\relax\fi
  \ifdim\caseremaining<\caseheight \caseheight=\caseremaining\fi
  \setbox\caseleftpart=\vsplit\caseleft to \caseheight
  \setbox\caserightpart=\vsplit\caseright to \caseheight
  \setbox\caseleftpart=\vbox{\unvbox\caseleftpart}
  \setbox\caserightpart=\vbox{\unvbox\caserightpart}
  \noindent\begin{minipage}[t]{\casewidth}
    \begin{tcolorbox}[enhanced,colback=basecolor,colframe=black!45,boxrule=0.4pt,
      before skip=0pt,after skip=0pt,left=6pt,right=6pt,top=5pt,bottom=5pt]
      \vbox to \caseheight{\unvbox\caseleftpart\vfil}
    \end{tcolorbox}
  \end{minipage}\hfill
  \begin{minipage}[t]{\casewidth}
    \begin{tcolorbox}[enhanced,colback=bncolor,colframe=black!45,boxrule=0.4pt,
      before skip=0pt,after skip=0pt,left=6pt,right=6pt,top=5pt,bottom=5pt]
      \vbox to \caseheight{\unvbox\caserightpart\vfil}
    \end{tcolorbox}
  \end{minipage}\par
  \ifvoid\caseleft
    \ifvoid\caseright\else\newpage\casepages\fi
  \else\newpage\casepages\fi}

\setbox\caseleft=\casecolumn{
\textbf{Base (11 doctor turns)}\par\smallskip\hrule\medskip
\textbf{Dr (1):} How often have you been vomiting after coughing, and are there any other symptoms accompanying the vomiting?\par\smallskip
\textbf{Pt:} It only happened once, actually. I don't have any other symptoms like a fever or headache, just the coughing and then the vomiting.\par\smallskip
\textbf{Dr (2):} Have you ever had episodes of coughing that led to vomiting before, and do you have any known medical conditions such as asthma or chronic lung disease?\par\smallskip
\textbf{Pt:} No, I don't think I've had this happen before. And no, I don't have asthma or any lung problems that I know of.\par\smallskip
\textbf{Dr (3):} Have you experienced any recent trauma or injury to your chest that might have caused a rib fracture?\par\smallskip
\textbf{Pt:} No, I haven't had any injuries or trauma to my chest recently.\par\smallskip
\par\smallskip{\color{black!60}\itshape [Doctor turns 4--6 omitted]}\par\smallskip

\textbf{Dr (7):} Have you experienced any episodes of severe coughing or coughing fits recently that might have led to vomiting?\par\smallskip
\textbf{Pt:} Yes, I've had some really bad coughing fits. They're pretty intense, and that's when I ended up throwing up.\par\smallskip
\textbf{Dr (8):} Can you describe the cough—how long has it been going on, is it productive, and have you experienced any chest pain or discomfort along with it?\par\smallskip
\textbf{Pt:} I don't have any chest pain or discomfort. I'm not sure exactly how long it's been going on, maybe a few days? And I don't think I'm coughing up any phlegm, it just feels like a dry, harsh cough.\par\smallskip
\par\smallskip{\color{black!60}\itshape [Doctor turn 9 omitted]}\par\smallskip

\textbf{Dr (10):} Have you experienced any heartburn, acid reflux, or nausea in addition to the coughing?\par\smallskip
\textbf{Pt:} No, I don't have any heartburn or acid reflux. I didn't feel nauseous until right after the coughing fit, when I threw up.\par\smallskip
\textbf{Dr (11):} Thank you for sharing your symptoms. Based on your history and the questions we've discussed, I believe we have gathered sufficient information to proceed with the next steps. Let's review what we've discussed and determine the best course of action.\par\smallskip
\medskip\textbf{Final diagnosis turn}\par\smallskip
\textbf{Reasoning:} The patient presents with a single episode of vomiting following intense coughing fits, with no prior history of similar events or underlying conditions like asthma or lung disease. The absence of chest trauma rules out rib fracture or pneumothorax. The lack of fever, headache, or other systemic symptoms suggests non-infectious causes. The cough is described as dry, harsh, and without productive sputum, pointing away from respiratory infections like bronchitis or pneumonia. The absence of stridor or respiratory distress rules out conditions like epiglottitis or croup. The lack of heartburn or acid reflux rules out GERD. Given the acute nature and the mechanism of injury (coughing leading to vomiting without respiratory compromise), the most likely diagnoses are spontaneous rib fracture (due to the sudden, intense coughing), bronchospasm/acute asthma exacerbation (as the cough could trigger bronchospasm despite no prior asthma history), bronchitis (as a dry cough may be a feature of viral bronchitis), pulmonary embolism (as sudden coughing fits could be associated with embolism), and acute laryngitis (due to the dry, harsh cough resembling acute laryngitis symptoms).\par\smallskip
\textbf{Differential:}
\begin{enumerate}[leftmargin=*,nosep,label=\arabic*.]
\item spontaneous rib fracture
\item bronchospasm / acute asthma exacerbation
\item bronchitis
\item pulmonary embolism
\item acute laryngitis
\end{enumerate}
}
\setbox\caseright=\casecolumn{
\textbf{\sys{} (11 doctor turns)}\par\smallskip\hrule\medskip
\par{\footnotesize\textit{Diagnostic Belief State: whooping cough (55.77\%); anaphylaxis (25.05\%); pneumonia (6.78\%); influenza (3.87\%); chronic rhinosinusitis (2.68\%).}}\par
\textbf{Dr (1):} Have you experienced any fever, sore throat, or difficulty breathing along with the vomiting after coughing?\par\smallskip
\textbf{Pt:} No, I don't have a fever, my throat doesn't hurt, and I'm not having trouble breathing. It's just the coughing that makes me throw up.\par\smallskip
\par{\footnotesize\textit{Diagnostic Belief State: whooping cough (48.36\%); anaphylaxis (32.10\%); chronic rhinosinusitis (6.87\%); gerd (3.52\%); pneumonia (2.03\%).}}\par
\textbf{Dr (2):} Have you had any recent changes in your diet or exposure to sick individuals?\par\smallskip
\textbf{Pt:} No, I haven't changed my diet. But I have been around someone who has whooping cough, so I'm worried I might have caught it.\par\smallskip
\par{\footnotesize\textit{Diagnostic Belief State: whooping cough (90.93\%); anaphylaxis (5.63\%); chronic rhinosinusitis (1.21\%); gerd (0.62\%); pneumonia (0.36\%).}}\par
\textbf{Dr (3):} How long has the cough lasting, and have you noticed any episodes of a distinctive 'whoop' sound when breathing in?\par\smallskip
\textbf{Pt:} I'm not sure exactly how long it's been going on, maybe a few days? And no, I haven't really noticed a specific 'whoop' sound, just the coughing fits that make me throw up.\par\smallskip
\par\smallskip{\color{black!60}\itshape [Doctor turn 4 omitted]}\par\smallskip

\par{\footnotesize\textit{Diagnostic Belief State: whooping cough (94.94\%); anaphylaxis (1.45\%); chronic rhinosinusitis (1.27\%); gerd (0.65\%); pneumonia (0.38\%).}}\par
\textbf{Dr (5):} Have you experienced post-tussive vomiting, where you vomit after coughing fits?\par\smallskip
\textbf{Pt:} Yes, that's exactly what's happening. The coughing fits are so bad that I end up throwing up right after.\par\smallskip
\par\smallskip{\color{black!60}\itshape [Doctor turns 6--9 omitted]}\par\smallskip

\par{\footnotesize\textit{Diagnostic Belief State: whooping cough (99.80\%); pneumonia (0.15\%); bronchitis (0.03\%); inguinal hernia (0.01\%); chronic rhinosinusitis (0.00\%).}}\par
\textbf{Dr (10):} Have you experienced any recent weight loss, night sweats, or fatigue that could indicate a chronic or systemic infection?\par\smallskip
\textbf{Pt:} No, I don't have any of those. I'm not losing weight, I don't sweat at night, and I don't feel tired or fatigued.\par\smallskip
\par{\footnotesize\textit{Diagnostic Belief State: whooping cough (99.93\%); bronchitis (0.03\%); pneumonia (0.03\%); inguinal hernia (0.01\%); anaphylaxis (0.00\%).}}\par
\textbf{Dr (11):} Thank you for sharing all these details. Based on your symptoms and history, I have gathered sufficient information to proceed with the next steps in your care. Have a good day!\par\smallskip
\medskip\textbf{Final diagnosis turn}\par\smallskip
\par{\footnotesize\textit{Diagnostic Belief State: whooping cough (99.93\%); bronchitis (0.03\%); pneumonia (0.03\%); inguinal hernia (0.01\%); anaphylaxis (0.00\%).}}\par
\textbf{Reasoning:} The patient presents with severe coughing fits leading to post-tussive vomiting. The history includes exposure to whooping cough, but no 'whoop' sound, fever, fatigue, weight loss, or night sweats. The key features are coughing and vomiting without systemic symptoms. Based on the Bayesian hints and the history, the top five diagnoses are: whooping cough (exposure and symptoms consistent), bronchitis (coughing fits), acute laryngitis (coughing and vomiting), acute rhinosinusitis (coughing may be associated), and urti (common cold with coughing fits). The patient denies fever or systemic symptoms, which may lower the likelihood of other infections like pneumonia.\par\smallskip
\textbf{Differential:}
\begin{enumerate}[leftmargin=*,nosep,label=\arabic*.]
\item whooping cough
\item bronchitis
\item acute laryngitis
\item acute rhinosinusitis
\item urti
\end{enumerate}
}
\casepages
\endgroup


\section{Prompts}
\label{app:prompts}

\subsection{Disease prior}
\begin{tcolorbox}[breakable,enhanced,
  colback=gray!5,colframe=gray!60,boxrule=0.4pt,
  left=6pt,right=6pt,top=4pt,bottom=4pt]
\begin{Verbatim}[fontsize=\small,breaklines=true,breakanywhere=true,breaksymbolleft={}]
<|system|> You are a medical epidemiology expert estimating population-level disease prevalence.

Provide your best-calibrated estimate using general medical and epidemiological knowledge.

Return the estimated number of current cases per 1,000,000 adults as an integer from 0 to 1,000,000. This representation is used to preserve precision for rare diseases. Do not round the estimate to a percentage or automatically return zero merely because the disease is rare.

<|user|>
Estimate the point prevalence of {disease_name} in the general adult population.

Interpret point prevalence as the number of randomly selected adults per 1,000,000 who currently have the disease.
\end{Verbatim}
\end{tcolorbox}

\subsection{Symptom given disease}
\begin{tcolorbox}[breakable,enhanced,
  colback=gray!5,colframe=gray!60,boxrule=0.4pt,
  left=6pt,right=6pt,top=4pt,bottom=4pt]
\begin{Verbatim}[fontsize=\small,breaklines=true,breakanywhere=true,breaksymbolleft={}]
<|system|> You are a medical expert estimating conditional symptom frequencies for a coarse-grained clinical ontology.

Estimate P(symptom | disease): the probability that a typical adult with the disease has the symptom during the active disease episode and would acknowledge it if specifically asked during an initial clinical evaluation.

Interpretation rules:
- Estimate symptom presence, not the probability that it is mentioned spontaneously.
- Consider the general adult patient population across typical clinical settings.
- If the disease label is broad, consider its standard clinical scope and average across its common subtypes and presentations according to their approximate frequency. Do not restrict it to a single subtype.
- If the symptom label is broad, include clinically equivalent ways in which that symptom may be experienced or described, but do not include unrelated manifestations.
- Estimate the probability attributable to the current disease episode, not the patient's lifetime probability of experiencing the symptom.
- Return a best-calibrated central estimate as an integer from 0 to 100.
- Use 0 or 100 only when the symptom is essentially never or almost always present, respectively.
- Perform the clinical reasoning internally and do not explain it.

<|user|>
Estimate P(symptom | disease) for the following pair.

Disease: {disease_name}
Symptom: {symptom_name}

What percentage of typical adult patients with this disease have this symptom during the active disease episode and would acknowledge it if specifically asked?
\end{Verbatim}
\end{tcolorbox}

\subsection{Disease given risk factor}
\begin{tcolorbox}[breakable,enhanced,
  colback=gray!5,colframe=gray!60,boxrule=0.4pt,
  left=6pt,right=6pt,top=4pt,bottom=4pt]
\begin{Verbatim}[fontsize=\small,breaklines=true,breakanywhere=true,breaksymbolleft={}]
<|system|> You are a medical epidemiology expert estimating conditional disease frequencies for a coarse-grained clinical ontology.

Estimate P(disease | risk factor): the probability that a randomly selected adult with the specified risk factor currently has the disease.

Interpretation rules:
- Estimate current disease status, not lifetime risk or the probability of developing the disease eventually.
- Treat the risk factor as present according to its standard clinical meaning.
- For time-dependent risk factors such as recent surgery, trauma, travel, exposure, or immobility, use the clinically relevant risk window for the disease.
- Average over the remaining characteristics of adults with the risk factor, including typical age, sex, comorbidities, and co-occurring risk factors.
- Do not interpret an association as certainty or assume that the risk factor is the sole cause of the disease.
- If the disease label is broad, average across its common subtypes according to their approximate frequency.
- If the risk-factor label is broad, use its usual population-level clinical interpretation rather than selecting an unusually severe form.
- Return the estimated number of current disease cases per 1,000,000 adults with the risk factor.
- Return a best-calibrated integer from 0 to 1,000,000.
- Do not automatically return zero merely because the disease is rare.
- Perform the medical reasoning internally and do not explain it.

<|user|>
Estimate P(disease | risk factor) for the following pair.

Risk factor: {risk_factor_name}
Disease: {disease_name}

Among 1,000,000 typical adults with this risk factor, approximately how many currently have this disease?
\end{Verbatim}
\end{tcolorbox}

\subsection{Base policy prompt}
\begin{tcolorbox}[breakable,enhanced,
  colback=gray!5,colframe=gray!60,boxrule=0.4pt,
  left=6pt,right=6pt,top=4pt,bottom=4pt]
\begin{Verbatim}[fontsize=\small,breaklines=true,breakanywhere=true,breaksymbolleft={}]
<|user|> Your task is to decide whether to **CONTINUE** the clinical interview with one meaningful next question or **END** the interview and conclude.

## **Available Information**
* **Possible Diagnoses:** A JSON list of diagnoses that define the allowable diagnostic space.
* **Conversation History:** The complete record of the interview so far.
* **Negative Constraints:** A list of questions that the doctor has already asked and has been answered. You are FORBIDDEN to use ask questions that are same, reworded or semantically similar to these questions. Use this list to avoid redundancy and ensure that each question explores a new clinical concept.

---

## **Interview Approach**

Conduct a focused, efficient diagnostic interview. Each question must have clear clinical value and reduce uncertainty.

* Start with a working differential based on the history so far and possible diagnoses.
* Ask only questions that could materially affect diagnostic reasoning.
* Do not repeat or rephrase questions targeting the same clinical concept.
* Treat explicit patient denials as valid and conclusive evidence.
* You are allowed a maximum of 60 questions in total.
* You must not end the interview before asking at least 5 questions.
* You are FORBIDDEN to ask questions that are already been asked and answered. The list of already asked questions is provided unde Negative Constraints.

---

## **Decision Logic**

Choose **END** when:
1. All clinically relevant domains suggested by the history and hints have been explored at least once.
2. Further questions would only revisit previously denied symptoms, medical history, family history, habits or exposures.
3. Additional information would be unlikely to change the diagnostic ranking.

Choose **CONTINUE** only if a single new question could plausibly alter the differential diagnosis.

---

## **Required Output**

Your output must be include three components in a strict JSON format:
1. reasoning: A detailed explanation of the reasoning leading to CONTINUE or END.
2. action: Either "CONTINUE" or "END".
3. response: The doctor's next single question if CONTINUE, or a closing statement if END.

Example output format:

```json
{
    "reasoning": "<Detailed explanation of the reasoning leading to CONTINUE or END>",
    "action": "CONTINUE or END",
    "response": "<Doctor's next single question OR closing statement>"
}

```

---

## **Input Data**

#### **Possible Diagnoses**

```json
{{diagnoses}}
```

#### **Number of Turns**
{{turns}}

#### **Conversation History**

{{history}}

#### **Negative Constraint**

You are FORBIDDEN from asking any of the following questions.
Do not ask them again in any form, including reworded, paraphrased, partially overlapping, or semantically equivalent questions.
These topics are already answered and must not be revisited.

{{constraints}}

\end{Verbatim}
\end{tcolorbox}

\subsection{Base differential diagnosis prompt}
\begin{tcolorbox}[breakable,enhanced,
  colback=gray!5,colframe=gray!60,boxrule=0.4pt,
  left=6pt,right=6pt,top=4pt,bottom=4pt]
\begin{Verbatim}[fontsize=\small,breaklines=true,breakanywhere=true,breaksymbolleft={}]
<|user|> You are an expert **Medical Doctor** tasked with evaluating the complete Conversation History to establish a final differential diagnosis.

## **Available Information**
* **Possible Diagnoses:** A JSON list of potential diagnoses. You MUST select exactly FIVE from this list.
* **Conversation History:** The complete and final record of the patient interview.

---

## **Clinical Approach**

Perform a focused synthesis of the patient's presentation using the Conversation History.

* Summarize key features: onset, duration, severity, associated symptoms, and pertinent negatives.
* The Possible Diagnoses list is the exclusive source of truth; do not include diagnoses outside this list.

CRITICAL: You MUST select diagnoses ONLY from the provided Possible Diagnoses list.
- Using any name not in this list is strictly forbidden.
- If a condition seems correct but is not in the list, you must choose the closest match or omit it.
- Failure to adhere will be considered an error.
- Always copy the name exactly as it appears in the list.

---

## **Differential Diagnosis Procedure**

1. **Clinical Synthesis:** Summarize the patient's presentation based on the conversation history.
2. **DDx Selection:** Select exactly FIVE diagnoses from the Possible Diagnoses list that best explain the patient's condition.
3. **Justification:** For each diagnosis, provide a concise explanation using only evidence from the history, including both positive findings and pertinent negatives.

---

## **Required Output**

Your output must be include three components in a strict JSON format:
1. reasoning: A detailed step-by-step explanation leading to the selection of the differential diagnosis.
2. response: A professional and empathetic closing statement summarizing the findings for the patient and introducing the differential diagnosis.
3. ddx: A list of exactly FIVE diagnoses from the provided list of potential diagnoses, each with a justification based on specific clinical evidence and denials from the patient's history

Example output format:

```json
{
  "action": "END",
  "reasoning": "<Detailed step-by-step explanation leading to the selection of the differential diagnosis>",
  "response": "<A professional and empathetic closing statement summarizing the findings for the patient and introducing the differential diagnosis>",
  "ddx": [
    {
      "diagnosis": "<Diagnosis name from the list>",
      "justification": "<Specific clinical evidence and denials from history supporting this diagnosis>"
    },
    {
      "diagnosis": "<Diagnosis name from the list>",
      "justification": "<Specific clinical evidence and denials from history supporting this diagnosis>"
    }
    // ... include FIVE total diagnoses
  ]
}

```

---

## **Input Data**

#### **Possible Diagnoses**

```json
{{diagnoses}}
```

#### **Number of Turns**
{{turns}}

#### **Conversation History**

{{history}}
\end{Verbatim}
\end{tcolorbox}

\subsection{Probmedtod policy prompt}
\begin{tcolorbox}[breakable,enhanced,
  colback=gray!5,colframe=gray!60,boxrule=0.4pt,
  left=6pt,right=6pt,top=4pt,bottom=4pt]
\begin{Verbatim}[fontsize=\small,breaklines=true,breakanywhere=true,breaksymbolleft={}]
<|user|> Your task is to decide whether to **CONTINUE** the clinical interview with one meaningful next question or **END** the interview and conclude.

## **Available Information**
* **Possible Diagnoses:** A JSON list of diagnoses that define the allowable diagnostic space.
* **Bayesian Network Hints:** A list of higher-probability conditions. Use these to guide your decision.
* **Conversation History:** The complete record of the interview so far.
* **Negative Constraints:** A list of questions that the doctor has already asked and has been answered. You are FORBIDDEN to use ask questions that are same, reworded or semantically similar to these questions. Use this list to avoid redundancy and ensure that each question explores a new clinical concept.

---

## **Interview Approach**

Conduct a focused, efficient diagnostic interview. Each question must have clear clinical value and reduce uncertainty.

* Ask only questions that could materially affect diagnostic reasoning.
* Do not repeat or rephrase questions targeting the same clinical concept.
* Treat explicit patient denials as valid and conclusive evidence.
* You are allowed a maximum of 60 questions in total.
* You must not end the interview before asking at least 5 questions.
* You are FORBIDDEN to ask questions that are already been asked and answered. The list of already asked questions is provided unde Negative Constraints.

---

## **Role of Bayesian Network Hints**

* Identify the most likely diagnoses from the hints. Use these to guide your decision.
* Prioritize questions that could confirm or rule out these high-probability conditions.
* If the Conversation History contradicts a hint, accept the contradiction and move on.

---

## **Decision Logic**

Choose **END** when:
1. All clinically relevant domains suggested by the history and hints have been explored at least once.
2. Further questions would only revisit previously denied symptoms or exposures.
3. Additional information would be unlikely to change the diagnostic ranking.

Choose **CONTINUE** only if a single new question could plausibly alter the differential diagnosis.

---

## **Required Output**

Your output must be include three components in a strict JSON format:
1. reasoning: A detailed explanation of the reasoning leading to CONTINUE or END.
2. action: Either "CONTINUE" or "END".
3. response: The doctor's next single question if CONTINUE, or a closing statement if END.

Example output format:

```json
{
    "reasoning": "<Detailed explanation of the reasoning leading to CONTINUE or END>",
    "action": "CONTINUE or END",
    "response": "<Doctor's next single question OR closing statement>"
}

```

---

## **Input Data**

#### **Possible Diagnoses**

```json
{{diagnoses}}
```

#### **Bayesian Network Hints**

```json
{{hints}}
```

#### **Number of Turns**
{{turns}}

#### **Conversation History**

{{history}}

#### **Negative Constraint**

You are FORBIDDEN from asking any of the following questions.
Do not ask them again in any form, including reworded, paraphrased, partially overlapping, or semantically equivalent questions.
These topics are already answered and must not be revisited.

{{constraints}}
\end{Verbatim}
\end{tcolorbox}

\subsection{Probmedtod differential diagnosis prompt}
\begin{tcolorbox}[breakable,enhanced,
  colback=gray!5,colframe=gray!60,boxrule=0.4pt,
  left=6pt,right=6pt,top=4pt,bottom=4pt]
\begin{Verbatim}[fontsize=\small,breaklines=true,breakanywhere=true,breaksymbolleft={}]
You are an expert **Medical Doctor** tasked with evaluating the complete Conversation History to establish a final differential diagnosis.

## **Available Information**
* **Possible Diagnoses:** A JSON list of potential diagnoses. You MUST select exactly FIVE from this list.
* **Bayesian Network Hints:** A list of higher-probability conditions. Use these to guide your decision.
* **Conversation History:** The complete and final record of the patient interview.

---

## **Clinical Approach**

Perform a focused synthesis of the patient's presentation using the Conversation History.

* Summarize key features: onset, duration, severity, associated symptoms, and pertinent negatives.
* Use the Bayesian Network Hints to prioritize likely diagnoses.
* If the history contradicts a hint, accept the contradiction and de-prioritize that hint.
* The Possible Diagnoses list is the exclusive source of truth; do not include diagnoses outside this list.

CRITICAL: You MUST select diagnoses ONLY from the provided Possible Diagnoses list.
- Using any name not in this list is strictly forbidden.
- If a condition seems correct but is not in the list, you must choose the closest match or omit it.
- Failure to adhere will be considered an error.
- Always copy the name exactly as it appears in the list.

---

## **Differential Diagnosis Procedure**

1. **Clinical Synthesis:** Summarize the patient's presentation based on the conversation history.
2. **DDx Selection:** Select exactly FIVE diagnoses from the Possible Diagnoses list that best explain the patient's condition.
3. **Justification:** For each diagnosis, provide a concise explanation using only evidence from the history, including both positive findings and pertinent negatives.

---

## **Required Output**

Your output must be include three components in a strict JSON format:
1. reasoning: A detailed step-by-step explanation leading to the selection of the differential diagnosis.
2. response: A professional and empathetic closing statement summarizing the findings for the patient and introducing the differential diagnosis.
3. ddx: A list of exactly FIVE diagnoses from the provided list of potential diagnoses, each with a justification based on specific clinical evidence and denials from the patient's history

Example output format:

```json
{
  "action": "END",
  "reasoning": "<Detailed step-by-step explanation leading to the selection of the differential diagnosis>",
  "response": "<A professional and empathetic closing statement summarizing the findings for the patient and introducing the differential diagnosis>",
  "ddx": [
    {
      "diagnosis": "<Diagnosis name from the list>",
      "justification": "<Specific clinical evidence and denials from history supporting this diagnosis>"
    },
    {
      "diagnosis": "<Diagnosis name from the list>",
      "justification": "<Specific clinical evidence and denials from history supporting this diagnosis>"
    }
    // ... include FIVE total diagnoses
  ]
}

```

---

## **Input Data**

#### **Possible Diagnoses**

```json
{{diagnoses}}
```

#### **Number of Turns**
{{turns}}

#### **Bayesian Network Hints**

```json
{{hints}}
```

#### **Conversation History**

{{history}}
\end{Verbatim}
\end{tcolorbox}

\subsection{Patient simulation prompt}\label{app:simulator_prompt}
\begin{tcolorbox}[breakable,enhanced,
  colback=gray!5,colframe=gray!60,boxrule=0.4pt,
  left=6pt,right=6pt,top=4pt,bottom=4pt]
\begin{Verbatim}[fontsize=\small,breaklines=true,breakanywhere=true,breaksymbolleft={}]
You are an AI model tasked with acting as a patient to help a doctor practice their clinical interview skills.
Your entire persona and medical history are defined by the information below.

**Your Goal:** To respond to the doctor's queries based *only* on the provided patient information and the ongoing conversation, convincingly portraying the persona of a typical patient.

---

### **Patient Profile**

You must strictly adhere to the following patient details. This is the **only** information you know about yourself.

{{note}}

---

### **Guiding Principles for Your Persona**

You must follow these rules without deviation to ensure a realistic interaction:

1.  **Stick to the Facts**: Base all your answers strictly on the **Patient Profile** above. Do not invent, add, or assume any information that is not explicitly written there. This is your source of truth.

2.  **Speak Like a Patient, Not a Textbook**: Use simple, common, and non-medical language.
    * **Describe Experiences**: Instead of just listing a diagnosis, describe the feeling or impact.
        * *Instead of:* "I have rheumatoid arthritis."
        * *Try:* "My joints get really stiff and painful, especially in my hands. The doctor called it arthritis."
    * **Use Simple Terms**: Describe pain as "sharp," "dull," or "achy." Refer to medications by their common name or purpose (e.g., "my water pill," "the little white one for my heart").
    * **Be a Little Vague**: Real patients often don't remember exact dates or clinical details. It's natural to say, "It started a few years ago, I think," or "I'm not sure of the name of the surgery, but it was on my knee."

3.  **Pace the Conversation (Avoid Information Dumps)**: This is crucial for a natural dialogue. Do not volunteer your entire medical history in one answer. Answer only the question asked, and wait for the doctor to ask follow-up questions.
    * If asked a broad question like, "Do you have any other medical conditions?", mention one or two of the most significant ones and let the doctor probe for more.
    * **Example:**
        * **Doctor:** "OK, and do you have any other medical conditions?"
        * **Correct Response:** "Well, I have high blood pressure, and my arthritis has been acting up for years. There are a few other things too."

4.  **Handle "Not Applicable" Information Correctly**: This rule applies to any inquiry not covered in your profile.
    * **This is critical:** If the doctor asks about any detail that is **NOT** in your profile, you must state that it is **not the case, that you don't have it, or you don't recall it.**
    * *Example (Symptom):* If your profile does **not** mention "headache" and the doctor asks, "Are you having any headaches?", you must answer, "**No, I don't have any headaches**."
    * *Example (Family History):* If your profile's family history only says "Father had high blood pressure" and the doctor asks, "Any family history of diabetes?", you must answer, "**No, not that I know of**."

5.  **Maintain Conversational Awareness**: Your response must **always** be a direct reaction to the doctor's immediately preceding statement. Acknowledge what the doctor says instead of just repeating old facts.
    * If the doctor acknowledges your problem and offers a solution, your response should be a natural reaction to their statement (e.g., gratitude, acknowledgment).
    * **Example:**
        * **Doctor:** "OK, so we'll make sure to get you some more of those [Lasix] today."
        * **Correct Response:** "**Oh, thank you, doctor. I'd appreciate that.**" or simply "**Okay, thank you.**"

6.  **Accept Guidance**: If the doctor explains what they think is wrong, gives a diagnosis, or outlines a plan, simply accept it politely. Do not argue or question their expertise.

7.  **Closing the Interview**: When the doctor indicates that they are finished, conclude the conversation by politely thanking them (e.g., "Thank you for your time, doctor.").

8.  **Be Concise**: Keep your responses to a reasonable length, ideally under 100 words. This helps maintain a natural flow of conversation and allows the doctor to ask follow-up questions.

### **Output Format**
You must respond a strict JSON object enclosed within ```json and ``` tags, following the structure shown below. Do not include any other text outside the JSON block.

```json
{
    "reasoning": "<Your internal reasoning about how you formulated your response>",
    "response": "<Your actual response to the doctor here>"
}

```

### **Conversation History**
Below is the conversation you have had with the doctor so far. The conversation hase {{turns}} turns. Use it to understand the context.

{{dialogue_history}}

---

### **Your Immediate Task**

Before you write your response, follow these steps:

1.  **Focus on the Final Turn:** Read the **very last thing the doctor said** in the `Conversation History`.
2.  **Determine the Doctor's Intent:** Is it a question? A reassurance? A statement of fact? A plan of action?
3.  **Formulate a Direct Reply:** Craft your response as a direct and logical answer or reaction to **that specific statement**, while strictly adhering to all the `Guiding Principles` above. Your answer must make sense as the immediate next line in the dialogue.

Provide your answer in the specified JSON format without any additional text.
\end{Verbatim}
\end{tcolorbox}

\section*{NeurIPS Paper Checklist}

\begin{enumerate}

\item {\bf Claims}
    \item[] Question: Do the main claims made in the abstract and introduction accurately reflect the paper's contributions and scope?
    \item[] Answer: \answerYes{} 
    \item[] Justification: We evaluate \sys{} on three datasets using a patient simulator. We report the results in section \ref{sec:results} showcasing performance gains.

\item {\bf Limitations}
    \item[] Question: Does the paper discuss the limitations of the work performed by the authors?
    \item[] Answer: \answerYes{} 
    \item[] Justification: We discuss limitations in \cref{app:limitations}, including the fixed disease set, the tripartite graph structure, the LLM used for BayesNet construction, and the monolingual evaluation.

\item {\bf Theory assumptions and proofs}
    \item[] Question: For each theoretical result, does the paper provide the full set of assumptions and a complete (and correct) proof?
    \item[] Answer: \answerNA{} 
    \item[] Justification: \sys{} does not include theoretical results, so this question is not applicable.

    \item {\bf Experimental result reproducibility}
    \item[] Question: Does the paper fully disclose all the information needed to reproduce the main experimental results of the paper to the extent that it affects the main claims and/or conclusions of the paper (regardless of whether the code and data are provided or not)?
    \item[] Answer: \answerYes{} 
    \item[] Justification: We provide detailed experimental setup in \cref{sec:expsetup}, including dataset details, patient simulator and evaluation protocol, and baselines. We also provide implementation details in \cref{app:implementation} and the full simulator prompt in \cref{app:simulator_prompt}. We include the code base in the suplemental material, and will release it publicly upon publication.

\item {\bf Open access to data and code}
    \item[] Question: Does the paper provide open access to the data and code, with sufficient instructions to faithfully reproduce the main experimental results, as described in supplemental material?
    \item[] Answer: \answerYes{} 
    \item[] Justification: Code and data are available at \url{https://github.com/dair-iitd/ProbMedTOD}. MIMIC-IV is publicly available post data use agreement, and DDXPlus is available.

\item {\bf Experimental setting/details}
    \item[] Question: Does the paper specify all the training and test details (e.g., data splits, hyperparameters, how they were chosen, type of optimizer) necessary to understand the results?
    \item[] Answer: \answerYes{} 
    \item[] Justification: Section \ref{sec:expsetup} provides detailed information on the experimental setup.

\item {\bf Experiment statistical significance}
    \item[] Question: Does the paper report error bars suitably and correctly defined or other appropriate information about the statistical significance of the experiments?
    \item[] Answer: \answerYes{} 
    \item[] Justification: We provide statistical significance tests for the main results in \cref{sec:results}.   

\item {\bf Experiments compute resources}
    \item[] Question: For each experiment, does the paper provide sufficient information on the computer resources (type of compute workers, memory, time of execution) needed to reproduce the experiments?
    \item[] Answer: \answerYes{} 
    \item[] Justification: Section \cref{app:implementation} provide detailed information on the compute resources.
    
\item {\bf Code of ethics}
    \item[] Question: Does the research conducted in the paper conform, in every respect, with the NeurIPS Code of Ethics \url{https://neurips.cc/public/EthicsGuidelines}?
    \item[] Answer: \answerYes{} 
    \item[] Justification:

\item {\bf Broader impacts}
    \item[] Question: Does the paper discuss both potential positive societal impacts and negative societal impacts of the work performed?
    \item[] Answer: \answerYes{} 
    \item[] Justification: We discuss the potential societal impacts of \sys{} in \cref{app:impact}.
    
\item {\bf Safeguards}
    \item[] Question: Does the paper describe safeguards that have been put in place for responsible release of data or models that have a high risk for misuse (e.g., pre-trained language models, image generators, or scraped datasets)?
    \item[] Answer: \answerNA{} 
    \item[] Justification:

\item {\bf Licenses for existing assets}
    \item[] Question: Are the creators or original owners of assets (e.g., code, data, models), used in the paper, properly credited and are the license and terms of use explicitly mentioned and properly respected?
    \item[] Answer: \answerYes{} 
    \item[] Justification: We cite the original papers for MIMIC-IV and DDXPlus.

\item {\bf New assets}
    \item[] Question: Are new assets introduced in the paper well documented and is the documentation provided alongside the assets?
    \item[] Answer: \answerNA{} 
    \item[] Justification:

\item {\bf Crowdsourcing and research with human subjects}
    \item[] Question: For crowdsourcing experiments and research with human subjects, does the paper include the full text of instructions given to participants and screenshots, if applicable, as well as details about compensation (if any)? 
    \item[] Answer: \answerNA{} 
    \item[] Justification:

\item {\bf Institutional review board (IRB) approvals or equivalent for research with human subjects}
    \item[] Question: Does the paper describe potential risks incurred by study participants, whether such risks were disclosed to the subjects, and whether Institutional Review Board (IRB) approvals (or an equivalent approval/review based on the requirements of your country or institution) were obtained?
    \item[] Answer: \answerNA{} 
    \item[] Justification:

\item {\bf Declaration of LLM usage}
    \item[] Question: Does the paper describe the usage of LLMs if it is an important, original, or non-standard component of the core methods in this research? Note that if the LLM is used only for writing, editing, or formatting purposes and does \emph{not} impact the core methodology, scientific rigor, or originality of the research, declaration is not required.
    \item[] Answer: \answerYes{} 
    \item[] Justification: We use LLMs for BayesNet construction, in \sys{} system and as a component in the patient simulator, which are important components of our work.

\end{enumerate}

\end{document}